\documentclass{article}
\usepackage[preprint,nonatbib]{neurips_2020}
\usepackage[utf8]{inputenc}
\usepackage[T1]{fontenc}
\usepackage{hyperref}
\usepackage{url}
\usepackage{booktabs}
\usepackage{amsfonts}
\usepackage{nicefrac}
\usepackage{microtype}
\usepackage{caption}
\usepackage{graphicx}
\usepackage{subfigure}
\usepackage{booktabs}
\usepackage{overpic}
\usepackage{commath}
\usepackage{bm}
\usepackage{verbatim}
\usepackage{xcolor}

\usepackage{amsmath,amsfonts,amssymb}
\usepackage{neurips_2020}
\usepackage{url}
\usepackage[export]{adjustbox}
\usepackage{wrapfig}

\newcommand{\myrefeq}[1]{Eq.~\eqref{#1}}
\newcommand{\myreffig}[1]{Fig.~\ref{#1}}

\newcommand{\myrefsec}[1]{Sec.~\ref{#1}}

\renewcommand{\v}[1]{\mathbf{#1}}
\definecolor{nilscol}{rgb}{0.8,0.15,0.08}
\newcommand{\you}[1]{{#1}}
\newcommand{\nilsE}[1]{{#1}}

\definecolor{stdCol}{rgb}{0.8,0.5,0.25}
\definecolor{ortCol}{rgb}{0.0,0.2,0.7}
\definecolor{rrCol}{rgb}{0.1,0.6,0.2}
\definecolor{rr1Col}{rgb}{0.56,0.93,0.56}
\definecolor{lrrCol}{rgb}{0.97,0.79,0.68}
\newcommand{\textStd}[1]{{\color{stdCol}#1}}
\newcommand{\textOrt}[1]{{\color{ortCol}#1}}
\newcommand{\textRr}[1]{{\color{rrCol}#1}}

\newcommand{\mrrPow}[2]{\text{RR}^{#2}_{\text{#1}}}
\newcommand{\mstd}[1]{\text{Std}_{\text{#1}}}

\newcommand{\vrrNosub}{{\color{rrCol}$\text{RR}$}}
\newcommand{\vrrPow}[2]{{\color{rrCol}$\mrrPow{#1}{#2}$}}
\newcommand{\vstd}[1]{{\color{stdCol}$\mstd{#1}$}}
\newcommand{\mort}[1]{\text{Ort}_{\text{#1}}}
\newcommand{\vort}[1]{{\color{ortCol}$\mort{#1}$}}
\newcommand{\data}[1]{\mathcal{D}_{#1}}
\newcommand{\bdata}[1]{D_{#1}}          
\newcommand{\datum}[1]{\v{d}_{#1}}       
\newcommand{\datumt}[2]{\v{d}^{#2}_{#1}} 
\newcommand{\Loss}[1]{\mathcal{L}_{#1}} 
\newcommand{\myspace}{\vspace{-2.4mm}}
\newcommand{\myspaceAfter}{\vspace{-1.5mm}}
\title{Data-driven Regularization via Racecar Training for Generalizing Neural Networks}

\author{
You Xie
\qquad \hspace{25pt}
Nils Thuerey \\
\hspace{14pt}{\tt you.xie@tum.de}
\hspace{14pt}{\tt nils.thuerey@tum.de} 
    \vspace{10pt} \\
Technical University of Munich\\
 https://github.com/tum-pbs/racecar
}

\let\OLDthebibliography\thebibliography
\renewcommand\thebibliography[1]{
	\OLDthebibliography{#1}
	\setlength{\parskip}{1pt}
	\setlength{\itemsep}{2pt plus 0.3ex}
}

\begin{document}

\maketitle

\begin{abstract}
  We propose a novel training approach for improving the generalization in neural networks. 
We show that in contrast to regular constraints for orthogonality,
our approach represents a {\em data-dependent} orthogonality constraint,
and is closely related to singular value decompositions of the weight matrices.
We also show how our formulation is easy to realize in practical network architectures 
via a reverse pass, which aims for reconstructing the full sequence of internal states of the network.
Despite being a surprisingly simple change, we demonstrate that this forward-backward training approach, which we refer to as {\em racecar} training, leads to significantly more generic features being extracted from a given data set.
Networks trained with our approach show more balanced mutual information
between input and output throughout all layers, 
yield improved explainability and,
exhibit improved performance for a variety of tasks and task transfers.
\end{abstract}

\myspace{}
\section{Introduction}
\myspaceAfter{}
Despite their vast success,
training neural networks that generalize well to a wide range of previously unseen
tasks remains a fundamental challenge \cite{neyshabur2017exploring,kawaguchi2017generalization,bansal2018can,frankle2018lottery}.
A variety of techniques have been proposed over the years, ranging from
well-established ones like dropout \cite{srivastava2014dropout}, and weight decay \cite{li2017input,hanson1989comparing,weigend1991generalization,krogh1992simple}, 
to techniques for orthogonalization \cite{ozay2016optimization,jia2017improving,bansal2018can}.
While the former ones primarily focus on avoiding over-fitting rather than generality, 
the latter is closer to the goals of our paper: generalization 
is often seen in terms of extracting
the most basic set of feature templates \cite{tang2013fundamental}, similar 
\begin{figure*}[h]
	\vspace{-1mm}
	\centering
	\includegraphics[width=0.99\linewidth]{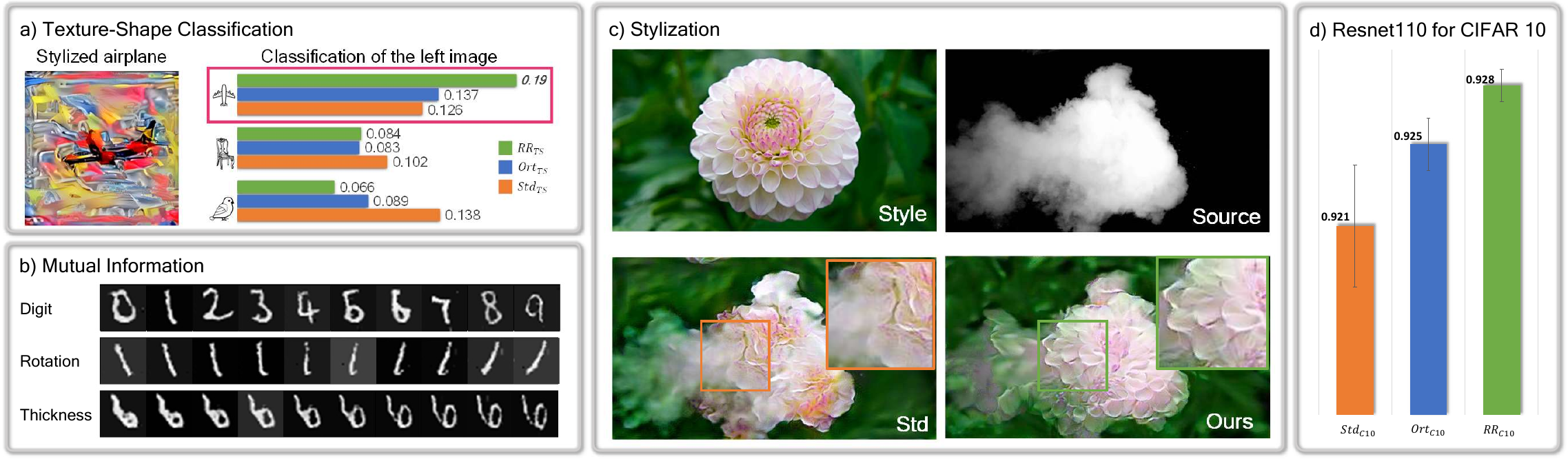}
	\vspace{-1mm}
	\caption{\footnotesize{
			Racecar training yields improved results for a wide range of applications:
			\textbf{a)}: For difficult shape classification tasks, our approach outperforms regular training (\vstd{TS}) as well as state-of-the-art regularizers (\vort{TS}):
			\nilsE{the \vrrPow{TS}{} model instead} classifies the airplane shape with high confidence.
			\textbf{b)}: 
			Our approach establishes mutual information between input and output distributions, 
			retrieving intuitive and explainable dimensions.        
			\nilsE{\textbf{c)}: A racecar model 
				yields an improved reconstruction for stylization tasks.
				\textbf{d)}: For CIFAR 10 classification with a Resnet110, 
				\vrrPow{C10}{} yields substantial practical improvements over the state-of-the-art \vort{C10}.}
	}} \label{fig:teaser}
	\vspace{-3mm}
\end{figure*}
to orthogonal vectors, that span the 
distribution of a data set under consideration.
Here, especially the dominant features \cite{phan2010tensor} that occur most often 
in a data set are those that have the highest potential 
to be applicable in new contexts and yield generalizing capabilities.
The challenge of extracting such features is two-fold:
finding the dominant features in a large and diverse data-set, 
and representing them given limited computational resources.
\nilsE{Our formulation incurs only a very moderate computational cost,
is very easy to integrate and widely applicable, while at the same time
outperforming existing methods for extracting dominant features.}

Our formulation draws inspiration from the human trait to disassemble and re-assemble objects.
These deconstruction tasks serve a wide range of purposes, from a hobby for motorists, to an 
essential source of learning and exploration for children \cite{gopnik1999scientist}.
We show that dominant, and general features can be learned with a 
simple, yet powerful and general modification of 
neural network training in line with these re-assembly tasks: in addition to a regular  
forward pass, we add a reverse pass that is constrained to reconstruct all in-between results of the forward pass as well as the input. This palindromic structure yields substantial improvements for generalization of the learned 
representations in a wide range of architectures and significantly improves performance.
Our results indicate that the reversible nature of the proposed training 
setup, which we will subsequently refer to via the palindrome {\em ''racecar''}
due to the backward-forward connectivity of the resulting network, 
encourages the learning of generic and reusable features that benefit a wide 
range of objectives. 

We will evaluate the generality of the extracted dominant features 
in the context of transfer learning. 
For a regular, i.e., a non-transfer task, the goal usually is to train a network that gives the optimal performance for  one specific goal. \nilsE{This has been demonstrated in numerous success stories \cite{lecun1998gradient, krizhevsky2012imagenet, goodfellow2014generative, he2016deep}. In a regular training run}, the network naturally exploits any observed correlations between input and output distribution. 
E.g., if the color of an object in any way correlates with its type, the training of a classifier should find and use this information. In recent years, even networks trained only for a very specific task were shown to be 
surprisingly suitable starting points when training models for different tasks \cite{zamir2018taskonomy,gopalakrishnan2017deep, ding2017facenet2expnet}. Often, the original network learned representations that were at least partially applicable to different data domains.
An inherent difficulty in this setting is that typically no knowledge about the specifics of the new data and task domains is available at the training time of the source model. While it is common practice to target broad and difficult tasks hoping that this will result in features that are applicable in new domains, 
we instead leverage the data distribution of the inputs when training the source model.

The structure of our networks is inspired
by invertible network architectures that have received significant attention
in recent years \cite{gomez2017reversible,jacobsen2018revnet,zhang2018rc}.
However, instead of aiming for a bijective mapping that reproduces inputs,
we strive for learning a general representation by constraining 
the network to represent an as-reversible-as-possible process 
for all {\em intermediate} layer activations.
Thus, even for cases where a classifier can, e.g., rely on color for inference of an object type, the model is encouraged to learn a representation that can recover the input. Hence, not only the color of the input should be retrieved, but also, e.g., its shape. 
In contrast to most structures for invertible networks,
racecar training does not impose architectural restrictions.
We demonstrate the benefits of racecar training for a variety of architectures, from 
fully connected layers to convolutional neural networks (CNNs), 
over networks with and without batch normalization, to GAN architectures.

\myspace{}
\section{Related Work}
\myspaceAfter{}
\nilsE{Several prior methods employed} "hard orthogonal constraints" to improve weight orthogonality via singular value decomposition (SVD) at training time \cite{huang2018orthogonal,jia2017improving,ozay2016optimization}.
Bansal et al.~\cite{bansal2018can} additionally investigated efficient formulations of the orthogonality constraints. 
In practice, these constraints are difficult to satisfy, and correspondingly only weakly imposed. 
Besides, 
these methods focus on improving performance for a known, given task. 
This means the training process only extracts features that the network considers useful for improving the performance of the current task, not necessarily improving generalization or 
transfer performance \cite{torrey2010transfer}. 
While our racecar training shares similarities with SVD-based constraints, it can be realized with a very efficient $L^2$-based formulation, and 
takes the full input distribution into account, leading to improved generalization.

Recovering all input information from hidden representations of a network is generally very difficult \cite{dinh2016density,mahendran2016visualizing}, due to the loss of information throughout the layer transformations.
In this context, \cite{tishby2015deep} proposed the information bottleneck principle, which states that for an optimal representation, information unrelated to the current task is omitted. This highlights the common specialization of conventional training approaches.
\nilsE{Reversed network architectures were proposed in previous work \cite{ardizzone2018analyzing,jacobsen2018revnet,gomez2017reversible}, }
but mainly focus on how to make a network fully invertible via augmenting the network with special structures. As a consequence, the path from input to output is different from the reverse path that translates output to input. 
Besides, the augmented structures of these approaches can be challenging to apply to general
network architectures. In contrast, our approach 
fully preserves an existing architecture for the backward path,
and does not require any operations that were not part of the source network. As such, 
it can easily be applied in new settings, e.g., adversarial training \cite{goodfellow2014generative}. 
While methods using reverse connections were previously proposed \cite{zhang2018rc, teng2019invertible}, 
these modules primarily focus on transferring information between layers 
for a given task, and on auto-encoder structures for domain adaptation, respectively.

Transfer learning with deep neural networks has been very successful for a variety of challenging tasks, such as image classification \cite{duan2012learning, kulis2011you, zhu2011heterogeneous}, multi-language text classification \cite{zhou2014heterogeneous, prettenhofer2010cross, zhou2014hybrid}, and medical imaging problems \cite{ravishankar2016understanding}. 
\cite{zamir2018taskonomy} proposed an approach to obtain task relationship graphs for different tasks. 
In this context, a central challenge is to set up the training process such that it leads to
learning generic features from the data set, which are useful for both source and related tasks 
rather than specific features for the source task.

\myspace{}
\section{Method}
\myspaceAfter{}
\label{sec:method}
Regularization via orthogonality has become a popular tool for training deep CNNs and was shown to improve performance \cite{huang2018orthogonal,bansal2018can}.
The corresponding constraints can be formulated as: 
\vspace{-2mm}
  \begin{equation}
  \setlength{\belowdisplayskip}{3pt}
\Loss{\text{ort}} = \sum_{m=1}^{n}\left \|M_{m}^{T} M_{m} - I\right \|_F^2 ,
 \label{eq:ortho_loss}
 \end{equation}
i.e., enforcing the transpose of the weight matrix $M_m\in \mathbb{R}^{s_{m}^{\text{out}}\times s_{m}^{\text{in}}}$ for all layers $m$ to yield its inverse when being multiplied with the original matrix. $I$ denotes the identity matrix with $I=(\v{e}_{m}^{1},...\v{e}_{m}^{s_{m}^{\text{in}}})$,  $\v{e}_{m}^{j}$ denoting the $j_{th}$ column unit vector.  
Minimizing \myrefeq{eq:ortho_loss}, i.e. $M_{m}^{T} M_{m} - I=0$
is mathematically equivalent to:
\vspace{-2mm}
 \begin{equation}
 M_{m}^{T} M_{m}\v{e}_{m}^{j} - \v{e}_{m}^{j}=\v0, j=1,2,...,s_{m}^{\text{in}},\\
 \label{eq:ortho_loss_2}
 \end{equation}
with $rank(M_{m}^{T}M_{m})=s_{m}^{\text{in}}$, and $\v{e}_{m}^{j}$ as eigenvectors of $M_{m}^{T}M_{m}$ with eigenvalues of 1. 
This formulation highlights that \myrefeq{eq:ortho_loss_2} does not depend on the training data, and instead only targets the content of $M_{m}$.
Our approach re-formulates the original orthogonality 
constraint in a {\em data-driven} manner: we take into account 
the set $\data{m}$ of inputs 
for the current layer (either activation from a previous layer or 
the training data $\data{1}$), and instead minimize 
\vspace{-2mm}
 \begin{equation}
\setlength{\belowdisplayskip}{2pt}
\Loss{\text{RR}}  = \sum_{m=1}^{n}(M_{m}^{T} M_{m}\datumt{m}{i} - \datumt{m}{i})^2 = \sum_{m=1}^{n}((M_{m}^{T} M_{m} - I) \datumt{m}{i})^2, \\
 \label{eq:rr_loss_0}
 \end{equation}
where $\datumt{m}{i} \in \data{m}  \subset \mathbb{R}^{s_{m}^{\text{in}}}$.
This re-formulation of orthogonality allows for minimizing the loss 
by extracting the dominant features of the input data
instead of only focusing on the content of $M_m$.

We use $q$ to denote the number of linearly independent entries in $\data{m}$, i.e. its dimension,
and $t$ the size of the training data, i.e. $|\data{m}|=t$, usually with $q<t$.
For every single datum $\datumt{m}{i}, i=1,2,...,t$,
\myrefeq{eq:rr_loss_0} results in 
 \begin{equation}
M_{m}^{T} M_{m} \datumt{m}{i} - \datumt{m}{i}=\v0,\\
 \label{eq:rr_loss_2}
 \end{equation}
and hence $\datumt{m}{i}$ are 
eigenvectors of $M_{m}^{T}M_{m}$ with corresponding eigenvalues being 1.
Thus, instead of the generic constraint $M_{m}^{T} M_{m}=I$ 
\nilsE{that is completely agnostic to the data at hand, 
the proposed formulation of \myrefeq{eq:rr_loss_2} is aware of the training data, which 
improves the generality of the learned representation, as we will demonstrate in detail below.}

As by construction, $rank(M_{m})=r\leqslant min(s_{m}^{\text{in}},s_{m}^{\text{out}})$, the SVD of $M_{m}$ yields:
\vspace{-1mm}
\begin{equation}
\begin{aligned}
M_{m}=U_{m} \Sigma_{m} V_{m}^{T}, \ \text{with} 
\left\{\begin{matrix}
U_{m}=(\v{u}_{m}^{1},\v{u}_{m}^{2},...,\v{u}_{m}^{r}, \v{u}_{m}^{r+1},...,\v{u}_{m}^{s_{m}^{\text{out}}})\in \mathbb{R}^{s_{m}^{\text{out}}\times s_{m}^{\text{out}}} , \\
V_{m}=(\v{v}_{m}^{1},\v{v}_{m}^{2},...,\v{v}_{m}^{r}, \v{v}_{m}^{r+1},...,\v{v}_{m}^{s_{m}^{\text{in}}})\in \mathbb{R}^{s_{m}^{\text{in}}\times s_{m}^{\text{in}}} ,
\end{matrix}\right.
 \end{aligned}
\label{eq:M_solution}
\end{equation}

with left and right singular vectors in $U_{m}$ and $V_{m}$, respectively, and $\Sigma_{m}$ having square roots of the $r$ eigenvalues of $M_{m}^{T} M_{m}$ on its diagonal. $\v{u}_{m}^{k}$ and $\v{v}_{m}^{k} (k=1,...,r)$ are the eigenvectors of $M_{m}M_{m}^{T}$ and $M_{m}^{T}M_{m}$, respectively \cite{wall2003singular}. 
Here, especially the right singular vectors in $V_{m}^{T}$ are 
important, as they determine which structures of the input are processed by the transformation $M_{m}$.
The original orthogonality constraint with \myrefeq{eq:ortho_loss_2} 
yields $r$ unit vectors $\v{e}_{m}^{j}$ as the eigenvectors of $M_{m}^{T}M_{m}$. 
Hence, 
the influence of \myrefeq{eq:ortho_loss_2} on $V_{m}$ is completely 
independent of training data and learning objectives.

\nilsE{Next, we shoe that $\Loss{\text{RR}}$ facilitates learning dominant features from a given data set.
For this, we consider an arbitrary basis to span the space of inputs }
$\data{m}$ for layer $m$.
Let $\mathcal{B}_{m}:\left \langle \v{w}_{m}^{1},...,\v{w}_{m}^{q}\right \rangle$ 
denote a set of $q$ orthonormal basis vectors obtained via a Gram-Schmidt process, with $t\!\geqslant\!q\!\geqslant\!r$,
and $\bdata{m}$ denoting the matrix of the vectors in $\mathcal{B}_{m}$.
As we show in more detail in the appendix, 
our constraint from \myrefeq{eq:rr_loss_2} requires eigenvectors of $M^{T}M$ to be $\v{w}_{m}^{i}$, 
with $V_{m}$ containing $r$ orthogonal vectors $(\v{v}_{m}^{1},\v{v}_{m}^{2},...,\v{v}_{m}^{r})$ from $\mathcal{D}_{m}$
 and $(s_{m}^{\text{in}}-r)$ vectors from the null space of $M$. 

We are especially interested in how $M_{m}$ changes w.r.t. input in terms of $\bdata{m}$, 
i.e., we express $\Loss{\text{RR}}$ in terms of $\bdata{m}$.
By construction, each input $\datumt{m}{i}$ can be represented 
as a linear combination via a vector of coefficients $\v{c}_{m}^{i}$
that multiplies $\bdata{m}$ so that $\datumt{m}{i}\!=\!\bdata{m}\v{c}_{m}^{i}$.
Since $M_{m} \datum{m}=U_{m} \Sigma_{m} V_{m}^{T}\datum{m}$, 
the loss $\Loss{\text{RR}}$ of layer $m$ can be rewritten as
  \begin{equation}
 \begin{aligned}
\Loss{\text{RR}_m}  &= (M_{m}^{T} M_{m}\datum{m} - \datum{m})^2=(V_{m} \Sigma_{m}^{T}\Sigma_{m} V_{m}^{T}\datum{m}- \datum{m})^2\\
&=(V_{m} \Sigma_{m}^{T}\Sigma_{m} V_{m}^{T}\bdata{m}\v{c}_{m}- \bdata{m}\v{c}_{m})^2,
 \end{aligned}
 \label{eq:rr_loss_m}
 \end{equation}
where we can assume that the coefficient vector $\v{c}_{m}$ is accumulated over the training data set size $t$ via $\v{c}_{m}=\sum_{i=1}^{t}\v{c}_{m}^{i}$, since eventually every single datum in $\data{m}$ will contribute to $\Loss{\text{RR}_{m}}$. 
The central component of \myrefeq{eq:rr_loss_m} is $V_{m}^{T}D_{m}$.
For a successful minimization, $V_{m}$ needs to retain those $\v{w}_{m}^{i}$
with the largest $\v{c}_{m}$ coefficients. As $V_{m}$ is typically severely limited
in terms of its representational capabilities by the number of adjustable weights in a network,
it needs to focus on the most important eigenvectors in terms of $\v{c}_{m}$ 
in order to establish a small distance to $\bdata{m}\v{c}_{m}$.
Thus, features
that appear multiple times in the input data with a corresponding factor in $\v{c}_{m}$ will 
more strongly contribute to minimizing $\Loss{\text{RR}_m}$.

To summarize, 
$V_{m}$ is driven towards containing $r$ orthogonal vectors $\v w_{m}^{i}$ 
that represent the most frequent features 
of the input data, i.e., the dominant features. 
Additionally, due to the column vectors of $V_{m}$ being mutually orthogonal, 
$M_{m}$ is encouraged to extract different features from the input. 
By the sake of being distinct and representative for the data set, these features 
have the potential to be useful for new inference tasks.
The feature vectors embedded in $M_m$ can be extracted from the network 
weights in practical settings, as we will demonstrate below.

\myspaceAfter{}
\paragraph{Realization in Neural Networks}
\begin{wrapfigure}{hR}{0.41\linewidth}
\vspace{-4mm}
	\begin{center}
		\includegraphics[width=0.9\linewidth]{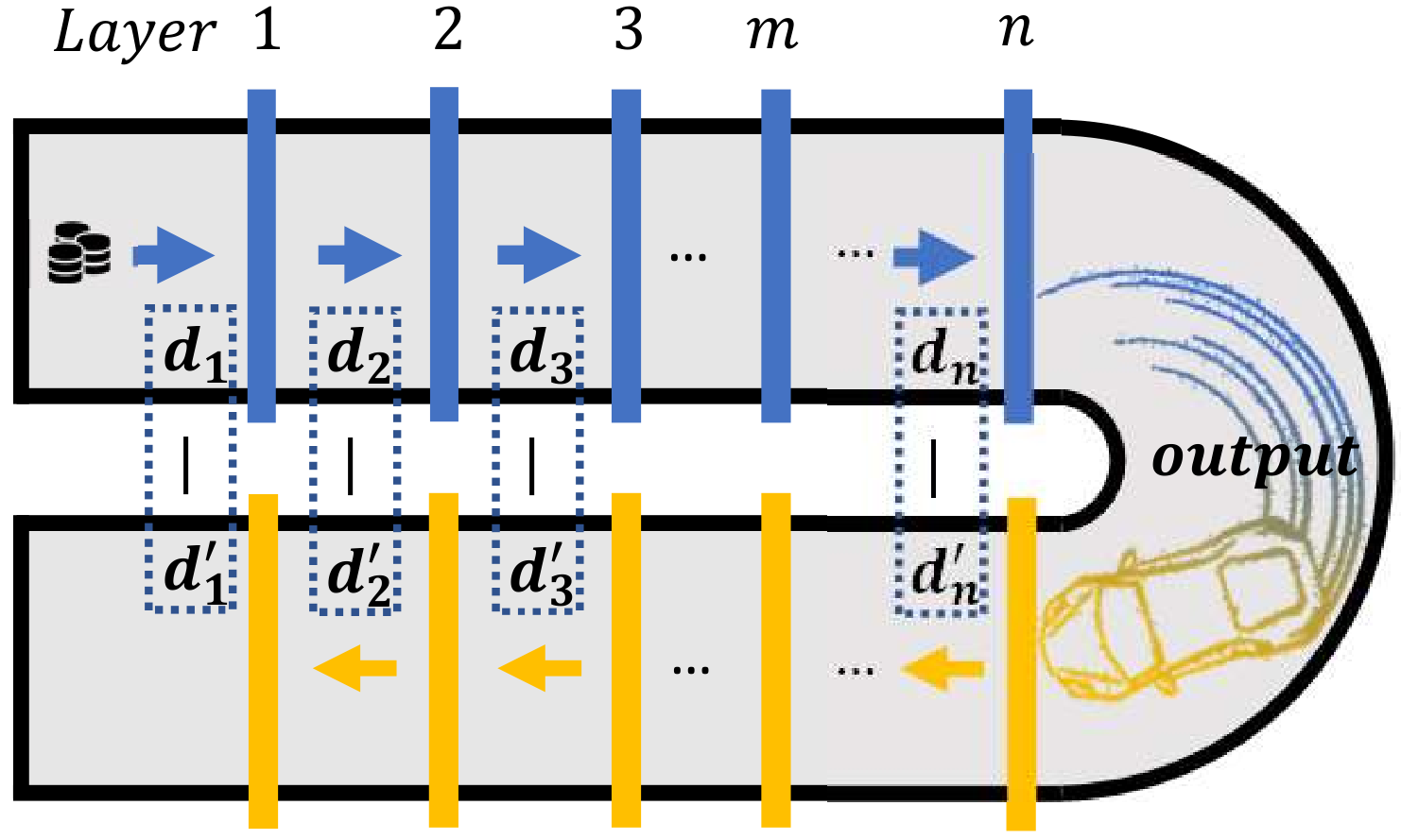}
	\end{center}
	\vspace{-4mm}
	\caption{\footnotesize{
	\nilsE{A visual overview of the regular forward pass (blue) and the corresponding 
  reverse pass (yellow) that we employ to learn via data-depend orthogonality constraints.} 
	} }
	\label{fig:pipeline}
	\vspace{-5mm}
\end{wrapfigure}
Calculating $M_{m}^{T} M_{m}$ is usually very expensive
due to the dimensionality of $M_{m}$. 
Instead of building it explicitly, we constrain intermediate results
to realize \myrefeq{eq:rr_loss_0} when training. 
\nilsE{A regular training typically starts} with a chosen network structure and trains the model weights for a given task via a suitable loss function. Our approach fully retains this setup and adds a second pass that reverses the initial structure while reusing all weights and biases. 
E.g., for a typical fully connected layer in the forward pass with $\datum{m+1} = M_{m} \datum{m} + \v{b}_{m}$, the reverse pass operation is given by $\datumt{m}{'} = M^{T}_{m} (\datum{m+1}-\v{b}_{m})$, where 
$\datumt{m}{'}$ denotes the reconstructed input.

Our goal with the reverse pass is to transpose all operations of the
forward pass to obtain identical intermediate activations between the layers with 
matching dimensionality. We can then constrain the  intermediate results of each 
layer of the forward pass to match the results of the backward pass, 
\nilsE{as illustrated in \myreffig{fig:pipeline}.}
Due to the symmetric structure of the two passes, 
we can use a simple $L^2$ difference to drive the network towards aligning the results:
\vspace{-3mm}
\begin{equation}
\setlength{\belowdisplayskip}{3pt}
\Loss{\text{racecar}}=\sum_{m=1}^{n}{\lambda_{m}\left \| \datum{m} - \datumt{m}{'}\right \|}_F^2.
\label{eq:loss}
\end{equation}
Here $\datum{m}$ denotes the input of layer $m$ in the forward pass and 
$\datumt{m}{'}$ the output of layer $m$ for the reverse pass. 
$\lambda_{m}$ denotes a scaling factor for the loss of layer $m$, which, however, is typically constant in our tests across all layers. Note that with our notation, $\datum{1}$ and  $\datum{1}^{'}$ refer to the input data, and the reconstructed input, respectively. 

Next, we show how this setup realizes the regularization from \myrefeq{eq:rr_loss_0}. 
For clarity, we use a fully connected layer with bias. In a neural network with $n$ hidden 
layers, the forward process for a layer $m$ is given by:
 \begin{equation}
  \setlength{\abovedisplayskip}{1pt}
 \setlength{\belowdisplayskip}{1pt}
 \datum{m+1}=M_{m} \datum{m}+\v{b}_{m}, 
 \label{eq:std_process}
 \end{equation}
 with $\datum{1}$ and $\datum{n+1}$ denoting in- and output, respectively.
For racecar training, we build a layer-wise reverse pass network with transposed operations and intermediate results $\datumt{m+1}{}$:
 \begin{equation}
 \setlength{\abovedisplayskip}{2pt}
 \setlength{\belowdisplayskip}{2pt}
 \datumt{m}{'}=M_{m}^{T}(\datumt{m+1}{}-\v{b}_{m}) ,
 \label{eq:rr_process}
 \end{equation}
\nilsE{ which yields ${\left \| \datum{m} - \datumt{m}{'}\right \|}_F^2=M_{m}^{T} M_{m} \datum{m} - \datum{m} $.}
\nilsE{When this difference is minimized via \myrefeq{eq:loss}, }
we obtain activated intermediate content 
during the reverse pass that reconstructs the values computed in the forward pass, i.e. 
$\datumt{m+1}{'}=\datum{m+1}$ holds. 
Replacing $\datum{m+1}$ with $\datum{m+1}{'}$ in \myrefeq{eq:rr_process}, 
yields a full reverse pass from output to input, which we use most racecar training runs below.
This version is preferable if a unique path from output to input exists. For architectures 
where the path is not unique, e.g., in the presence of \nilsE{additive residual connections}, 
we use the layer-wise formulation from \myrefeq{eq:rr_process}.
The full formulation instead gives:
 \begin{equation}
  \setlength{\abovedisplayskip}{2pt}
 \setlength{\belowdisplayskip}{2pt}
 \begin{aligned}
 \datumt{m}{'}=M_{m}^{T}(\datumt{m+1}{'}-\v{b}_{m})
 &=M_{m}^{T}(\datum{m+1}-\v{b}_{m})
 =M_{m}^{T}M_{m}\datum{m} \ , 
 \end{aligned}
 \label{eq:rr_process_2}
 \end{equation}
which is consistent with \myrefeq{eq:rr_loss_0}, and hence satisfies the 
original constraint $M_{m}^{T} M_{m} \datum{m} - \datum{m}=\v0$. 
Up to now, the discussion focused on simplified neural networks without activation functions or extensions such as batch normalization (BN). 
While we leave incorporating such extensions into the derivation for future work, our experiments consistently show that the inherent properties of racecar training remain valid: even with activations and BN, our approach successfully extracts dominant structures and yields improved generalization. In the appendix, we give details on how to ensure that the latent space content for forward and reverse pass is aligned such that differences can be minimized.

To summarize, we realize the loss formulation of \myrefeq{eq:loss} 
to minimize $\sum_{m=1}^{n}((M_{m}^{T} M_{m} - I) \datum{m})^2$ 
without explicitly having to construct $M_{m}^{T} M_{m}$.
\nilsE{We will refer to networks trained with the added reverse structure and the additional loss 
terms 
as being trained with {\em racecar training}.
We consider two variants for the reverse pass: a layer-wise racecar training \myrefeq{eq:rr_process}
using the local datum $\datum{m+1}$, and a full version via \myrefeq{eq:rr_process_2} which 
uses $\datumt{m+1}{'}$.}

\myspaceAfter{}
\paragraph{Experiments}
\begin{wrapfigure}{hR}{0.43\linewidth}
    \vspace{-4mm}
    \centering
        \includegraphics[width=1.0\linewidth]{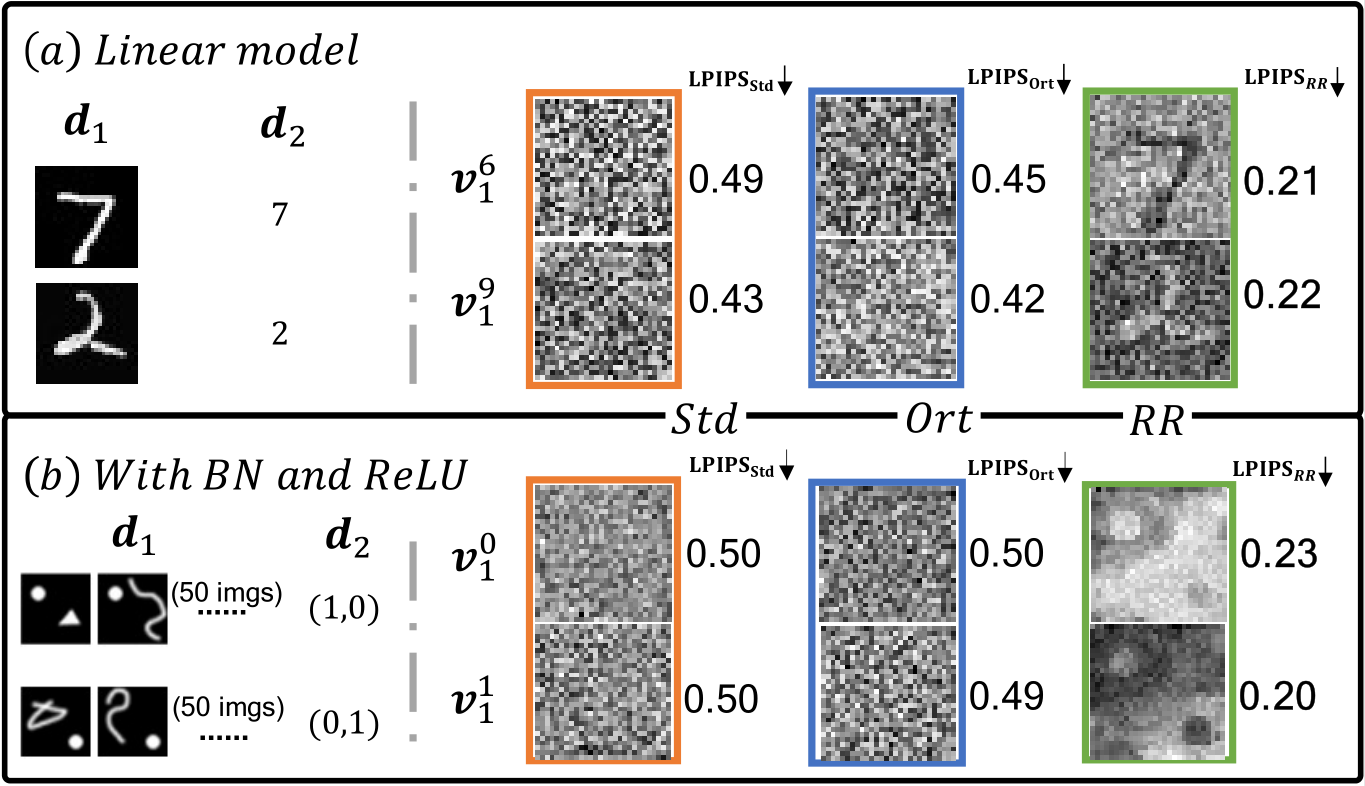}
    \vspace{-4mm}
    \caption{\footnotesize{Column vectors of $V_{m}$ for different trained models \vstd{}, \vort{} and \vrrPow{}{}: (a) MNIST and (b) for peaks. Input features clearly are successfully embedded in the weights of \vrrPow{}{}. }
    } \label{fig:SVD}
    \vspace{-3mm}
\end{wrapfigure}
We will use the following naming scheme to differentiate variants in the experiments below: 
\vstd{T}, \vrrPow{T}{}, \vort{T} for base models trained for a task T. 
Here, \vstd{} denotes a regular training run (in \textStd{orange color} in graphs below), while 
\vrrNosub$^{\color{rrCol}{}}$ 
denotes models trained with our racecar training (\textRr{in green}).
While we typically use all layers of a network in the racecar constraints, 
a special reduced variant that we compare to below only applies the constraint
for the input data, i.e., m=1 in our notation.
A network trained with this variant, denoted by 
{\color{rr1Col}$\text{RR}_\text{A}^{1}$},
is effectively trained to only reconstruct the input. \nilsE{It contains  
no constraints for the inner activations of the network.}
\textOrt{$\mort{}$} additionally denotes models trained with Spectral Restricted Isometry Property (SRIP) 
orthogonal regularization \cite{bansal2018can} (\textOrt{in blue}). 
We verify that the column vectors of $V_{m}$ of models from racecar training contain the dominant features of the input with the help of two classification tests. Both cases employ a single hidden fully connected layer, i.e. $\datum{2}=M_{1} \datum{1}$. In the first MNIST test, the training data consists only of 2 different images. 
After training, we compute the SVD for $M_{1}$.
Qualitatively, 
\vrrPow{}{} shows $v_{m}^{i}$ that have obvious similarity with the two inputs, 
while the vectors of \vstd{} and \vort{} contain no recognizable structures (\myreffig{fig:SVD}a).
To quantify this similarity, we compute an LPIPS distance \cite{zhang2018unreasonable} between $v_{m}^{i}$ and the training data (lower values being better).
These measurements confirm that the singular values of the \vrrPow{}{} model with $0.282 \pm 0.82$ are significantly more similar to the input images that \vstd{} and \vort{}, with $0.411 \pm 0.05$ and $0.403 \pm 0.031$, respectively.

For the second classification test,  
we employ a training data set that is constructed from two dominant classes (a peak in the top left, and bottom right quadrant, respectively), augmented with noise in the form of random scribbles. Based on the analysis above, we expect the racecar training to extract the two dominant peaks during training. \nilsE{Our results confirm this: the peaks are clearly visible for racecar training, an example is shown in \myreffig{fig:SVD}(b),} while the other models fail to extract structures that resemble the input. The LPIPS distances confirm the visual similarity with $0.217 \pm 0.022$ for \vrrPow{}{}, $0.500 \pm 0.002$ for \vstd{} and $0.495 \pm 0.006$ for \vort{}. 
\nilsE{Thus, our training approach yields a way for humans to inspect 
the structures learned by a neural network.} 

The results above experimentally confirm our derivation of the racecar loss, and its ability to extract dominant and  generalizing structures from the training data.
We have focused on relatively simple setups here, as this allows us to quantify similarity via the training data. 
Our results below indicate that racecar training similarly extracts dominant structures from the activations
for the deeper layers of a network. However, due to the lack of ground truth data, it \nilsE{is not easy} 
to evaluate these structures. Thus, we turn to measurements in terms of mutual information in the  next section.

The constraints of \myrefeq{eq:loss} intentionally only minimize differences in an averaged manner with an $L^2$ norm, as we don't strive for a perfectly bijective mapping between in- and output domains. Instead, racecar training aims at encouraging the
network to extract dominant features that preserve as much information from the input data set as possible, given the representative capabilities of the chosen architecture. Hence, while regular training allows a model to specialize its extracted features to maximize performance for a given task, 
our approach considers the full data distribution of the input \nilsE{to represent its dominant features.}

\myspace{}
\section{Evaluation in Terms of Mutual Information}
\myspaceAfter{}
\label{sec:MItest}
As our approach hinges on the introduction of the reverse pass, we will show that it succeeds in terms of establishing mutual information (MI) between the input and the constrained intermediates inside a network.
More formally, MI $I(X; Y)$ of
random variables $X$ and $Y$ measures how different the joint distribution of $X$ and $Y$ is w.r.t. the product of their marginal distributions, i.e., the Kullback-Leibler divergence 
$I(X;Y) = D_{KL}[P_{(X,Y)}||P_X P_Y]$.
\cite{tishby2015deep} proposed {\em MI plane} to analyze 
trained models, which show the MI between the input $X$ and 
activations of a layer $\data{m}$, i.e., $I(X;\data{m})$ and $I(\data{m};Y)$, i.e., MI of layer $\data{m}$ with output $Y$.
These two quantities indicate how much information about the in- and output distributions are retained at each layer, and we use them to show to which extent racecar training succeeds at incorporating information about the inputs throughout training. 

\begin{figure*}
    \vspace{-2mm}
    \centering
        \begin{overpic}[width=1.0\linewidth]{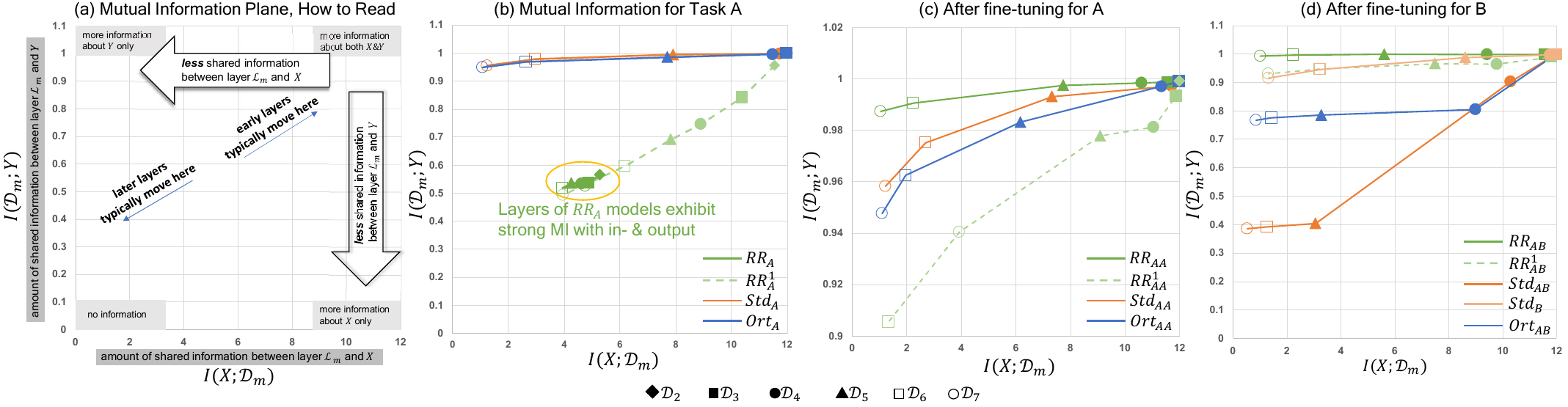}
        \end{overpic}
    \vspace{-2mm}
    \caption{\footnotesize{
    MI plane comparisons for different models. a) Visual overview of the content. b) Plane of task A. 
    Points on each line correspond to layers of one type of model.
    All points of \vrrPow{A}{}, 
    are located in the center of the graph, while \vstd{A} and \vort{A}, exhibit large $I(\data{m};Y)$, i.e., specialize on the output. 
    c,d): After fine-tuning for A/B. 
    The last layer $\data{7}$ of \vrrPow{AA}{} builds the strongest relationship with $Y$.
    } } \label{fig:informationplane}
    \vspace{-6mm}
\end{figure*} 

\begin{wrapfigure}{hR}{0.45\linewidth}
    \vspace{-5mm}
	\begin{center}
	    \includegraphics[width=1.0\linewidth]{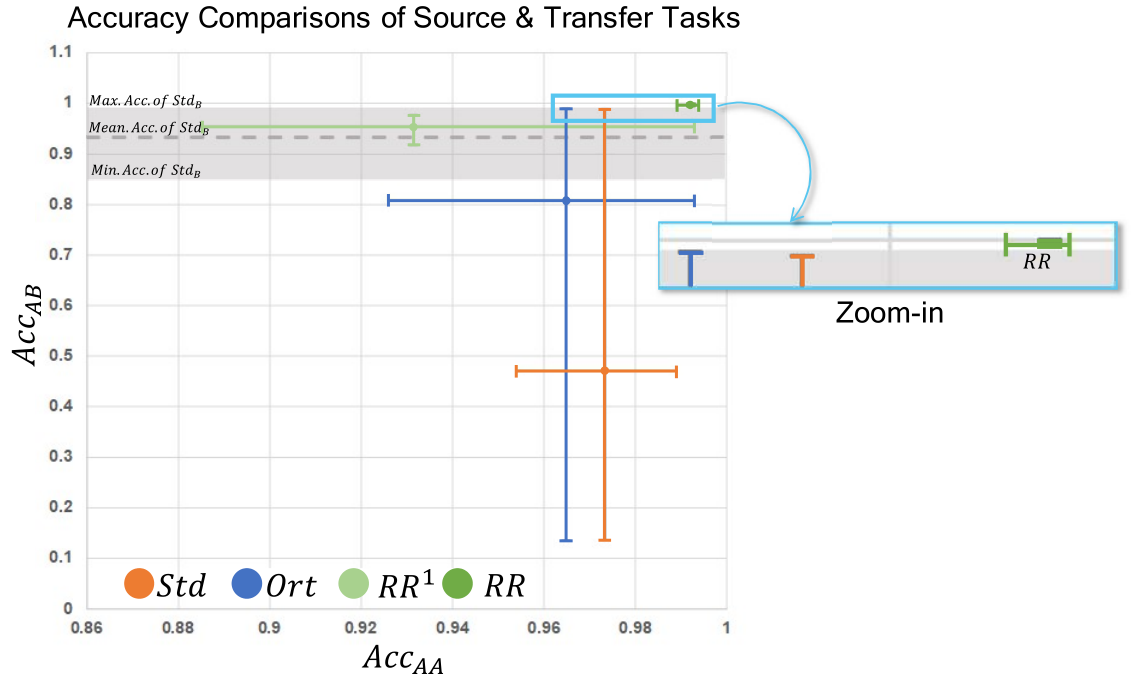}
	\end{center}
	\vspace{-4mm}
	\caption{\footnotesize{Performance for MI source and transfer tasks for the models of \myreffig{fig:informationplane}. Due to the large standard deviation of \vort, we show min/max value ranges.
	The dashed gray line and region show baseline accuracy for \vstd{B}. The top-left inset highlights the stability of the high accuracy results from racecar training.
	}}
	\label{fig:informationplane_accuracy}
	\vspace{-3mm}
\end{wrapfigure}
The following tests employ networks with six fully connected layers with the objective to learn the mapping from 12 binary inputs to 2 binary output digits \cite{shwartz2017opening}, with results accumulated over five runs. 
We compare \nilsE{the versions \vstd{A}, \vort{A}, \vrrPow{A}{},
and a variant of the latter:  {\color{rr1Col}$\text{RR}_\text{A}^{1}$}, i.e. 
a version} where only the input $\datum{1}$ is constrained to be reconstructed. 
While \myreffig{fig:informationplane} (a), visually summarizes the content of the MI planes, the graph in (b) highlights that racecar training correlates input and output distributions across all layers: the cluster of green points in the center of the graph shows that all layers contain balanced MI between in- as well as output and the activations of each layer. 
{\color{rr1Col}$\text{RR}_\text{A}^{1}$} fares slightly worse, while \vstd{A} and \vort{A} almost exclusively focus on the output with $I(\data{m};Y)$ being close to one.
Once we continue fine-tuning these models without regularization, the MI naturally shifts towards the output, as shown in \myreffig{fig:informationplane} (c). 
\vrrPow{AA}{} outperforms others in terms of final performance. Likewise, \vrrPow{AB}{}  performs  best for a transfer task B with switched output digits, as shown in graph (d).
The final performance for both tasks across all runs is summarized in \myreffig{fig:informationplane_accuracy}. 
\nilsE{It is apparent that our approach gives the highest performance and very stable results.}
These graphs visualize that racecar training succeeds in robustly extracting reusable features.

MI has received attention recently as a learning objective, e.g., in the form of the InfoGAN approach  \cite{chen2016infogan} for learning disentangled and interpretable latent representations.
Interestingly, a side-effect of the racecar training is that it successfully establishes mutual information between inputs and outputs, as shown in the previous paragraphs. And while MI is typically difficult to assess and estimate \cite{walters2009estimation}, our approach provides a simple and robust way for including it as a learning objective. In this way, we can, e.g., 
\nilsE{reproduce the disentangling results from \cite{chen2016infogan},} 
as shown in \myreffig{fig:teaser}(c).  
A generative model with racecar training extracts intuitive latent dimensions for the different digits, line thickness, and orientation without any additional modifications of the loss function.

\myspace{}
\section{Experimental Results}
\myspaceAfter{}
\label{sec:experiments}
To illustrate that racecar training is a practical approach that can be employed in real-world deep learning scenarios, we now turn to a broad range of complex network structures, i.e., CNNs, Autoencoders, and GANs, with a variety of data sets and tasks. 
We provide quantitative and qualitative results to show our approach succeeds in improving the generality of neural networks. 

\myspaceAfter{}
\paragraph{Transfer-learning Benchmarks}
\begin{wrapfigure}{hR}{0.40\linewidth}
    \vspace{-5mm}
    \centering
    \includegraphics[width=0.99\linewidth]{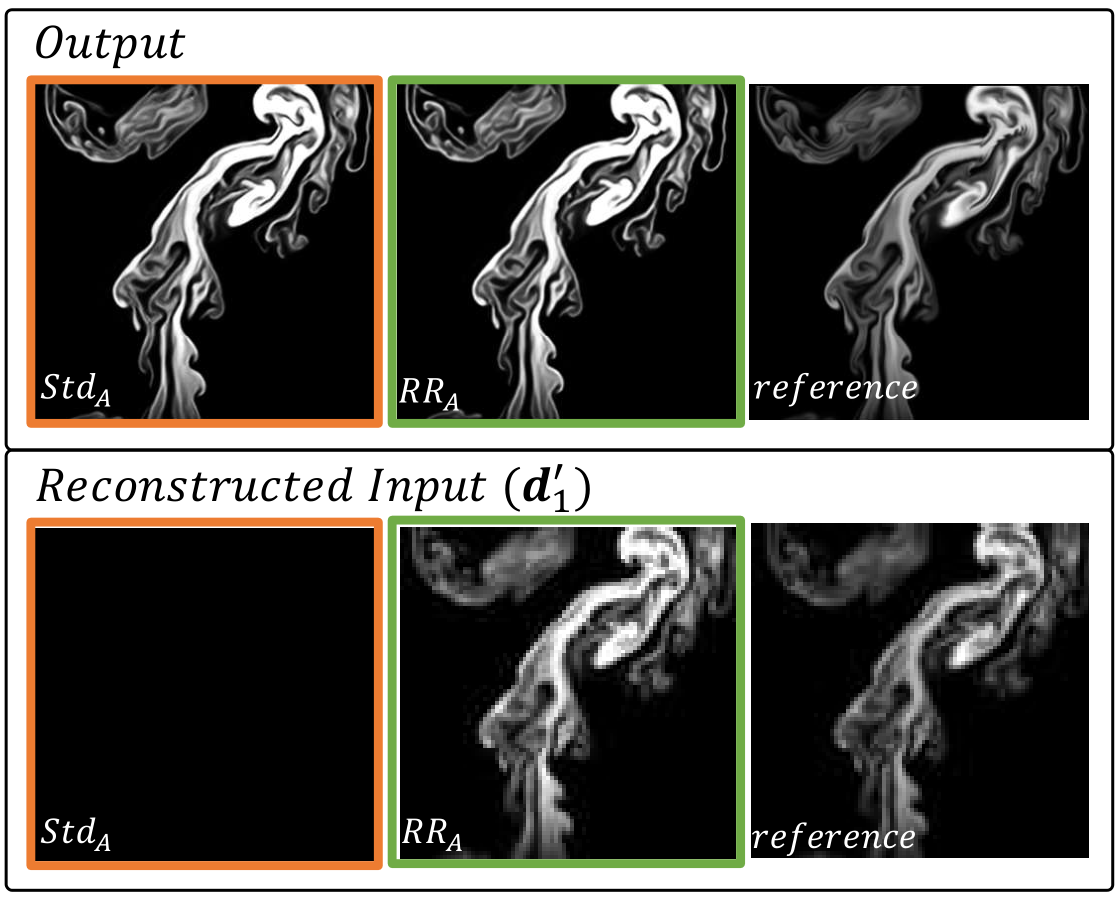}
    \vspace{-6mm}
    \caption{\footnotesize{
    Example output and reconstructed low resolution results, 
    with the reference shown right. 
    Only \vrrPow{A}{} successfully recovers the input, \vstd{A} produces a black image.} 
    } \label{fig:SmoHigh}
    \vspace{-2mm}
\end{wrapfigure}
We first evaluate our approach with two state-of-the-art transfer learning benchmark data sets. The first one uses the texture-shape data set from \cite{geirhos2018imagenet}, which contains challenging images of various shapes combined with patterns and textures to be classified. 
The results below are given for 10 runs each. 
For the stylized data shown in \myreffig{fig:texture_accuracy} (a), \vstd{TS} yields a performance of 44.2\%, and \vort{TS} improves the performance to 47.0\%, while \vrrPow{TS}{} yields a performance of 54.7\% (see  \myreffig{fig:texture_accuracy}b).
Thus, the accuracy of \vrrPow{TS}{} is $23.76\%$ higher than \vstd{TS}, and $16.38\%$ higher than \vort{TS}. 
To assess generality, we also apply the models to new data without retraining, i.e. an edge and a filled 
data set, also shown in \myreffig{fig:texture_accuracy} (a). For the edge data set, \vrrPow{TS}{} outperforms 
both \vstd{TS} and \vort{TS} by $50\%$ and $16.75\%$, respectively, 
\begin{figure}
\vspace{-4mm}
\begin{minipage}{.49\linewidth}
  \centering
  \includegraphics[height=1.25in]{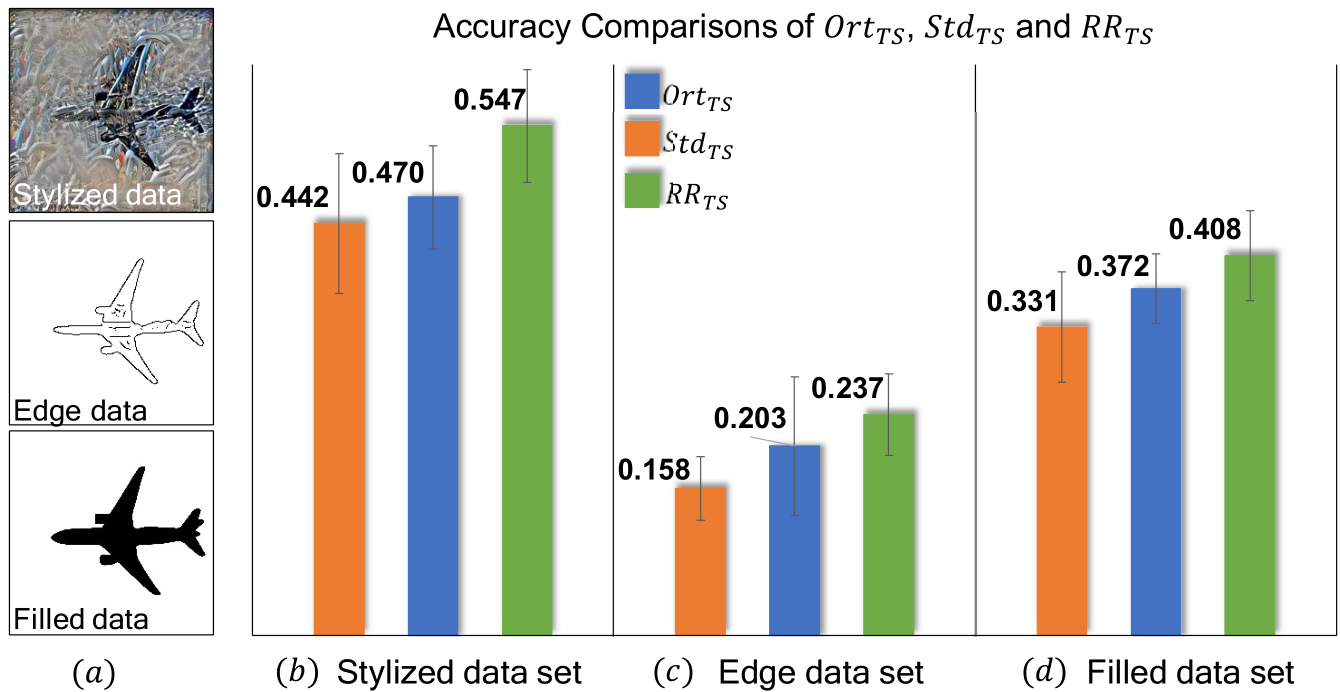}
  \captionsetup{width=.94\textwidth}
  \caption{\footnotesize{
     (a) Examples from texture-shape data set.
     (b, c, d) Texture-shape test accuracy comparisons of \vort{TS}, \vstd{TS} and \vrrPow{TS}{} for different data sets.}}
  \label{fig:texture_accuracy}
\end{minipage}
\begin{minipage}{.49\linewidth}
  \centering
  \includegraphics[height=1.25in]{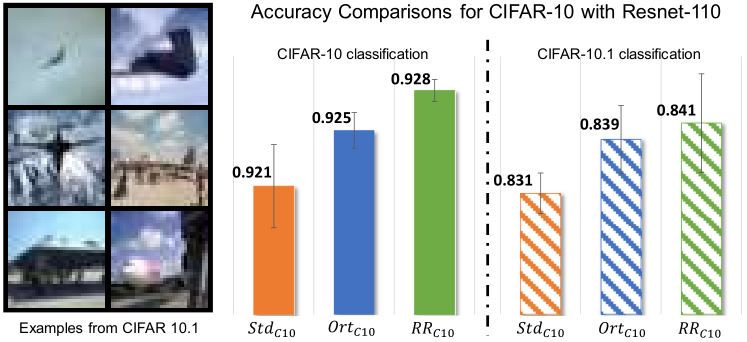}
  \captionsetup{width=.94\textwidth}
  \caption{\footnotesize{
     Left: Examples from CIFAR 10.1 data set. Right:  Accuracy comparisons when applying models trained on CIFAR 10 to CIFAR 10.1 data. }}
  \label{fig:cifar10_10.1}
\end{minipage}
\label{fig:fig}
\vspace{-3mm}
\end{figure}
It is worth pointing out that the additional constraints of our racecar training lead to increased 
requirements for memory and additional computations, e.g., 
41.86\% more time per epoch than regular training for the texture-shape test.
On the other hand, it allows us to train smaller models: 
we can reduce the weight count by 32\%
for the texture-shape case while still being on-par with \vort{TS} in terms of classification performance. 

As a second test case, we use a CIFAR-based task transfer \cite{recht2019imagenet} that measures how well models trained on the original CIFAR 10, generalize to a new data set (CIFAR 10.1) collected according to the same principles as the original one. 
Here we use a Resnet110 with 110 layers and 1.7 million parameters,  
In terms of accuracy across 5 runs, \vort{C10} outperforms \vstd{C10} by 0.39\%, 
while \vrrPow{C10}{} outperforms \vort{C10} by another 0.28\% in terms of absolute test accuracy (\myreffig{fig:cifar10_10.1}).
\nilsE{This increase for racecar training matches the gains reported for orthogonality in previous work \cite{bansal2018can},
and thus shows that our approach yields substantial practical improvements over the latter.}
It is especially interesting how well performance for CIFAR 10 translates into transfer performance for CIFAR 10.1. 
Here, \vrrPow{C10}{} still outperforms \vort{C10} and \vstd{C10} by 0.22\% and 0.95\%, respectively. 
Hence, the models from racecar training very successfully translate gains in performance from the original 
task to the new one, which indicates that the models have successfully learned a set of more general features.
To summarize, both benchmark cases confirm that racecar training benefits generalization. 
As racecar training consistently outperforms orthogonalization,  
we focus on comparisons with regular training in the following. 

\begin{wrapfigure}{hR}{0.38\linewidth}
    \vspace{-5mm}
    \centering
    \includegraphics[width=0.99\linewidth]{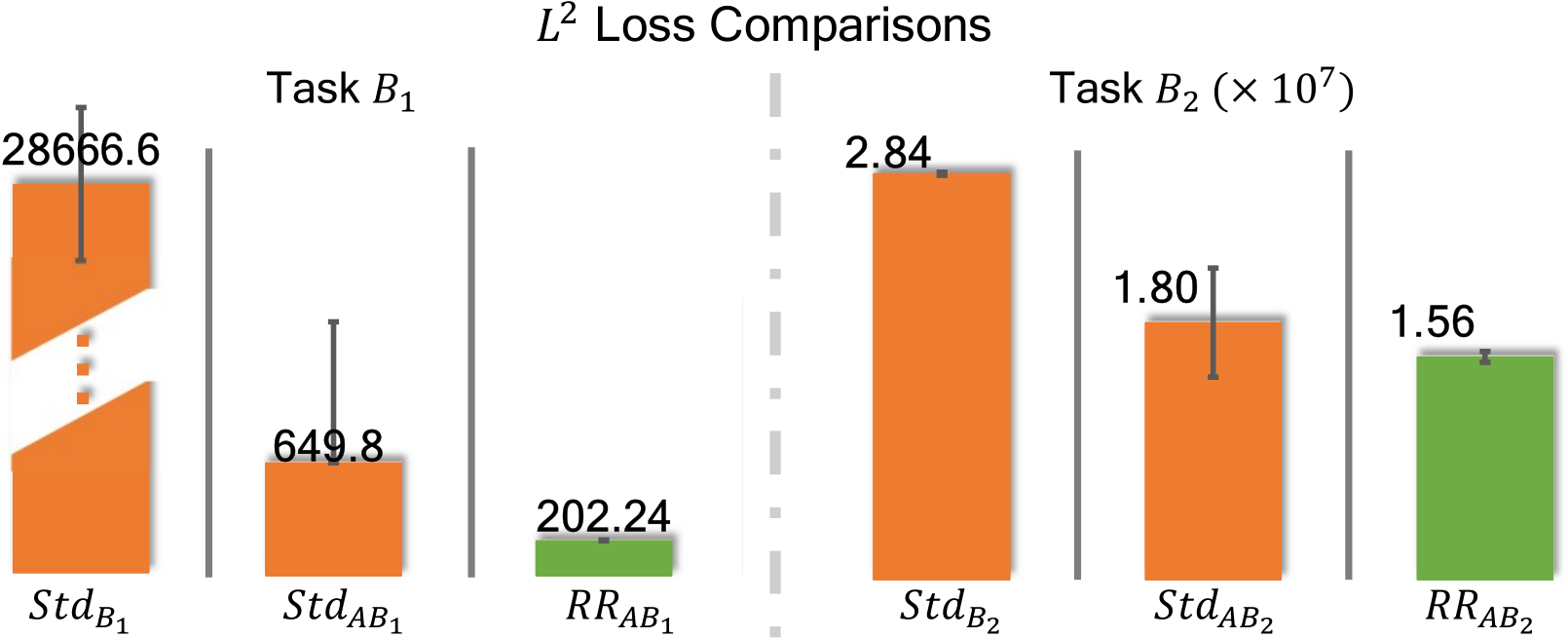}
    \vspace{-6mm}
    \caption{\footnotesize{$L^2$ loss comparisons for two different smoke transfer learning tasks (averaged across 5 runs each).
    The \vrrPow{}{} models show the best performance for both tasks.}
    } \label{fig:simAEbar}
    \vspace{-2mm}
\end{wrapfigure}

\myspaceAfter{}
\paragraph{Generative Adversarial Models}
In this section, we employ racecar training in the context of generative 
models for transferring from synthetic to real-world data from the ScalarFlow 
data set \cite{eckert2019scalarflow}. 
As \nilsE{super-resolution} task $A$, we first use a fully-convolutional 
generator network, adversarially trained with a discriminator network on the synthetic flow data.
\myreffig{fig:SmoHigh} demonstrates that the racecar network structure works in conjunction with 
the GAN training. As shown in the bottom row, the trained generator succeeds in recovering the input 
via the reverse pass without modifications. 
\nilsE{A regular model \vstd{A}, only yields a black image in this case.}

We now mirror the generator model from the previous task to create an autoencoder structure that we apply to two different data sets: the synthetic smoke data used for the GAN training (task $B_1$), and a real-world RGB data set of smoke clouds (task $B_2$). 
Thus both variants represent transfer tasks, the second one being more difficult due to the changed data distribution.
The resulting losses, summarized in \myreffig{fig:simAEbar},
show that racecar training performs best for both autoencoder tasks: the $L^2$ loss of \vrrPow{$AB_{1}$}{} is $68.88\%$ lower than \vstd{$AB_{1}$} for $B_1$, and $13.3\%$ lower for task $B_2$.
The latter is especially encouraging, as it represents a transfer from training with fully synthetic images to real-world images.

\myspaceAfter{}
\paragraph{VGG19 Stylization}
\begin{wrapfigure}{hR}{0.43\linewidth}
    \vspace{-6mm}
	\begin{center}
		\includegraphics[width=0.99\linewidth]{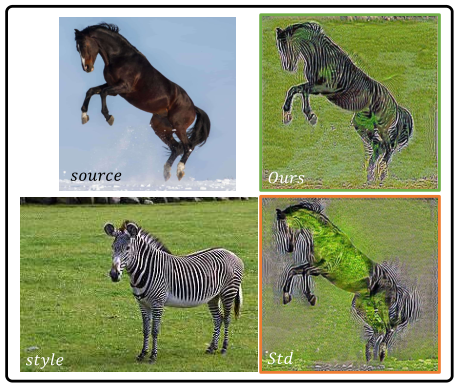}
	\end{center}
	\vspace{-3mm}
	\caption{\footnotesize{
	Stylization cases from horse to zebra.
	}}
	\label{fig:vgg2}
	\vspace{-6mm}
\end{wrapfigure}
To provide a qualitative evaluation in a complex visual scenario, we turn to image stylization. 
We use VGG$_{19}$ networks \cite{simonyan2014very} \nilsE{with more than 142 million parameters \cite{deng2009imagenet},
and train models with and without racecar training, i.e. \vrrPow{}{} and \vstd{}.
Both networks are then used for stylization tasks \cite{gatys2016image}. 
As this task strongly depends on the features of the base network, it
visualizes the differences of the learned structures.}

Several results of such stylizations are shown in \myreffig{fig:teaser} and \myreffig{fig:vgg2}. While the former highlights improved reconstruction of the leaf structures of the flower, 
the latter shows the popular setup of transforming a horse into a zebra. 
In \myreffig{fig:vgg2}, the regular model has difficulties distinguishing 
the background and the horse, and inserts green patches into the body, In contrast, the VGG$_{19}$ network 
with racecar training yields a significantly better output.
While these images only provide qualitative information about the properties of 
racecar training due to the lack of a ground truth result, we have found across a wide range of tests 
that performing stylization with models from racecar training yields more intuitive and robust results.
This is most likely caused by the improved representation these networks form of the 
input data distribution, which allows them to more clearly separate salient objects and visual styles.

\myspace{}
\section{Conclusions}
\myspaceAfter{}
We have proposed a novel training approach for improving neural network generalization by adding a constrained reverse pass. 
We have shown for a wide range of scenarios, from singular value decompositions, over mutual information, to transfer learning benchmarks, that racecar training yields networks with better generalizing capabilities.
Our training approach is very general and imposes no requirements regarding network structure or training methods. 
As future work, we believe it will be very interesting to evaluate our approach in additional contexts, e.g., 
\nilsE{for temporal structures \cite{hochreiter1997lstm,cho2014gru}, }
and for training explainable and interpretable models \cite{zeiler2014visualizing,chen2016infogan,du2018techniques}.

\clearpage

\newcommand{\mytoptitlebar}{
  \hrule height 2pt
  \vskip 0.25in
  \vskip -\parskip%
}
\newcommand{\mybottomtitlebar}{
  \vskip 0.29in
  \vskip -\parskip
  \hrule height 1pt
  \vskip 0.09in%
}
\mytoptitlebar
{\centering
{\Large Supplemental Material for {\em Data-driven Regularization via Racecar Training for Generalizing Neural Networks} \par}}
\mybottomtitlebar

\vspace{15pt}

\appendix

Below, we provide additional details regarding derivation, data sets, network architectures, and experiments mentioned in the main paper. 
We additionally show results that could not be included in the main document due to page restrictions:
an MNIST classification case (\myrefsec{sec:MNISTtestapp}), natural image classification (\myrefsec{sec:Cifartest}), and additional 
generation and stylization results in \myrefsec{sec:Smoketestapp} and \myrefsec{sec:addStyle}, respectively.

To ensure reproducibility, source code for all tests will be published, in addition to all data sets that are not readily available.
The runtimes below were measured on a machine with Nvidia GeForce GTX 1080 Ti GPUs and an Intel Core i7-6850K CPU.

\section{Method}
\label{sec:method}

\subsection{Racecar Loss and SVD}
\label{sec:gradient}

\begin{figure}[b!]
    \centering
    \includegraphics[width=0.99\linewidth]{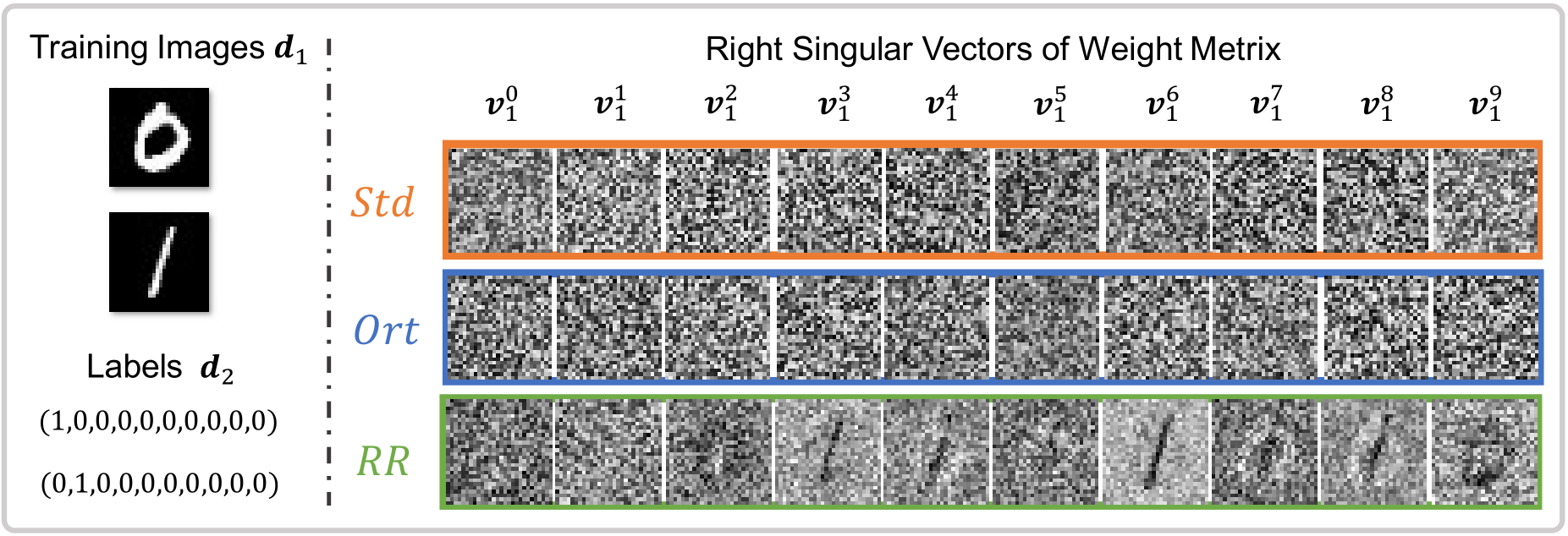}
    \caption{\footnotesize{
        An example of the intuitive analysis of network weights enabled by racecar training:
        we show the singular vectors for a conventional model \vstd{}, a model trained with orthogonal constraints \vort{}, and a racecar version \vrrPow{}{}. The singular vectors from racecar training show clear structures that resemble the inputs, while 
        the other variants contain no visible structures. 
        The evaluation via LPIPS of $\datum{1}$ versus all $\v{v}_{1}$ confirms this: it yields 
        0.316 for \vrrPow{}{} on aveage, which is significantly better than \vort{} with 0.504) and \vstd{} with 0.484.
        Details and further exampels are given in \myrefsec{sec:Blobtestapp} below.
    }}
    \label{fig:teaser}
    \vspace{-4mm}
\end{figure}

\nilsE{In this section we give a more detailed derivation of our racecar loss }
formulation and extends on the information given in Section 3 of the main paper.
As explained there, the racecar loss aims for minimizing 
\begin{equation}
\begin{aligned}
\Loss{\text{RR}}  = \sum_{m=1}^{n}(M_{m}^{T} M_{m}\datumt{m}{i} - \datumt{m}{i})^2 , 
\end{aligned}
\label{eq:rr_loss_o}
\end{equation}
where $M_m\in \mathbb{R}^{s_{m}^{\text{out}}\times s_{m}^{\text{in}}}$ denotes the weight matrix of layer $m$, and 
data from the input data set $\data{m}$ 
is denoted by $\datumt{m}{i} \subset \mathbb{R}^{s_{m}^{\text{in}}}, i=1,2,...,t$.
\nilsE{Here t denotes the number of samples in the input data set.}
Minimizing \myrefeq{eq:rr_loss_o} is mathematically equivalent to 
\begin{equation}
\begin{aligned}
M_{m}^{T} M_{m}\datumt{m}{i} - \datumt{m}{i}=\v0
\end{aligned}
\label{eq:rr_loss_1}
\end{equation}
for all $\datumt{m}{i}$.
Hence, perfectly fulfilling \myrefeq{eq:rr_loss_o} would require all $\datumt{m}{i}$ to be eigenvectors of $M_{m}^{T}M_{m}$ with corresponding eigenvalues being 1.
As in Sec. 3 of the main paper, we make use of an auxiliary	orthonormal basis $\mathcal{B}_{m}:\left \langle \v{w}_{m}^{1},...,\v{w}_{m}^{q}\right \rangle$, for which $q$ (with $q \leq t$) denotes the number of linearly independent entries in $\data{m}$.
While $\mathcal{B}_{m}$ never has to be explicitly constructed for our method,
\nilsE{it can, e.g., be obtained via Gram-Schmidt.
The matrix consisting of the vectors in $\mathcal{B}_{m}$ is denoted by $\bdata{m}$.}

Since the $\v{w}_{m}^{h} (h=1,2,...q)$ necessarily can be expressed as linear combinations of $\datumt{m}{i}$, 
\myrefeq{eq:rr_loss_o} similarly requires $\v{w}_{m}^{h}$ to be eigenvectors of $M_{m}^{T}M_{m}$ with corresponding eigenvalues being 1, i.e.:
\begin{equation}
\begin{aligned}
M_{m}^{T} M_{m}\v{w}_{m}^{h} - \v{w}_{m}^{h}=\v0
\end{aligned}
\label{eq:rr_loss_2}
\end{equation}	
We denote the vector of coefficients to express
$\datumt{m}{i}$ via $\bdata{m}$ with $\v{c}_{m}^{i}$, i.e.
$\datumt{m}{i}\!=\!\bdata{m}\v{c}_{m}^{i}$.
Then \myrefeq{eq:rr_loss_1} can be rewritten as:
\begin{equation}
\begin{aligned}
M_{m}^{T} M_{m}\bdata{m}\v{c}_{m}^{i} - \bdata{m}\v{c}_{m}^{i}=\v0
\end{aligned}
\label{eq:rr_loss_3}
\end{equation}	
Via an SVD of the matrix $M_m$ in \myrefeq{eq:rr_loss_3} we obtain 
\begin{equation}
\begin{aligned}
&M_{m}^{T} M_{m}\bdata{m}\v{c}_{m} - \bdata{m}\v{c}_{m}\\
&=\sum_{h=1}^{q}M_{m}^{T} M_{m}\v{w}_{m}^{h}\v{c}_{m_h} - \v{w}_{m}^{h}\v{c}_{m_h} \\
&=\sum_{h=1}^{q}V_{m} \Sigma_{m}^{T}\Sigma_{m} V_{m}^{T}\v{w}_{m}^{h}\v{c}_{m_h}-\v{w}_{m}^{h}\v{c}_{m_h}
\end{aligned}
\label{eq:rr_loss_5}
\end{equation}
where the coefficient vector $\v{c}_{m}$ is accumulated over the training data set size $t$ via $\v{c}_{m}=\sum_{i=1}^{t}\v{c}_{m}^{i}$. Here we assume that over the course of a typical training run eventually every single datum in $\data{m}$ will contribute to $\Loss{\text{RR}_{m}}$. 
This form of the loss highlights that minimizing $\Loss{\text{RR}}$ requires an alignment of $V_{m} \Sigma_{m}^{T}\Sigma_{m} V_{m}^{T}\v{w}_{m}^{h}\v{c}_{m_h}$ and $\v{w}_{m}^{h}\v{c}_{m_h}$.

By construction, $\Sigma_{m}$ contains the square roots of the 
eigenvalues of $M_{m}^{T} M_{m}$ as its diagonal entries.
The matrix has rank $r=rank(M_{m}^{T}M_{m})$, and since
all eigenvalues are required to be 1 by \myrefeq{eq:rr_loss_2}, 
the multiplication with $\Sigma_{m}$ in \myrefeq{eq:rr_loss_5} effectively 
performs a selection of $r$ column vectors from $V_{m}$.
Hence, we can focus on the interaction between the basis vectors $\v{w}_{m}$ and the $r$ active column vectors of $V_{m}$:
\begin{equation}
\begin{aligned}
& V_{m} \Sigma_{m}^{T}\Sigma_{m} V_{m}^{T}\v{w}_{m}^{h}\v{c}_{m_h} - \v{w}_{m}^{h}\v{c}_{m_h} \\
&=\v{c}_{m_h}(V_{m} \Sigma_{m}^{T}\Sigma_{m} V_{m}^{T}\v{w}_{m}^{h} - \v{w}_{m}^{h})\\
& =\v{c}_{m_h}(\sum_{f=1}^{r}(\v{v}_{m}^{f})^{T}\v{w}_{m}^{h}\v{v}_{m}^{f}-\v{w}_{m}^{h}).
\end{aligned}
\label{eq:rr_loss_6}
\end{equation}
As $V_{m}$ is obtained via an SVD it contains $r$ orthogonal eigenvectors of $M_{m}^{T}M_{m}$. \myrefeq{eq:rr_loss_2} requires  $\v{w}_{m}^{1},...,\v{w}_{m}^{q}$ to be eigenvectors of $M_{m}^{T}M_{m}$, 
but since typically the dimension of the \nilsE{input data set} is much larger than the dimension of the weight matrix, i.e. $r \leq q$, 
in practice only $r$ vectors from $\mathcal{B}_{m}$ can fulfill \myrefeq{eq:rr_loss_2}. This means the vectors $\v{v}_{m}^{1},...,\v{v}_{m}^{r}$ in $V_{m}$ are a subset of the orthonormal basis vectors  $\mathcal{B}_{m}:\left \langle \v{w}_{m}^{1},...,\v{w}_{m}^{q}\right \rangle$ with $(\v{w}_{m}^{h})^{2}=1$. 
Then for any $\v{w}_{m}^{h}$  we have
\begin{equation}
\begin{aligned}
\left\{\begin{matrix}
(\v{v}_{m}^{f})^{T}\v{w}_{m}^{h}=1,& \text{ if } \hspace{0.3mm} \v{v}_{m}^{f}=\v{w}_{m}^{h} \\ 
(\v{v}_{m}^{f})^{T}\v{w}_{m}^{h}=0,& \text{ otherwise}.
\end{matrix}\right.
\end{aligned}
\label{eq:rr_loss_9}
\end{equation}

Thus if $V_{m}$ contains $\v{w}_{m}^{h}$, we have 
\begin{equation}
\begin{aligned}
\sum_{f=1}^{r}(\v{v}_{m}^{f})^{T}\v{w}_{m}^{h}\v{v}_{m}^{f}=\v{w}_{m}^{h},
\end{aligned}
\label{eq:rr_loss_10}
\end{equation}
and we trivially fulfill the constraint 
\begin{equation}
\begin{aligned}
\v{c}_{m_h}(\sum_{f=1}^{r}(\v{v}_{m}^{f})^{T}\v{w}_{m}^{h}\v{v}_{m}^{f}-\v{w}_{m}^{h})=\v0.
\end{aligned}
\label{eq:rr_loss_7}
\end{equation}
However, 
due to $r$ being smaller than $q$ in practice,
$V_{m}$ typically can not include all vectors from $\mathcal{B}_{m}$. 
Thus, if $V_{m}$ does not contain $\v{w}_{m}^{h}$, we have  $(\v{v}_{m}^{f})^{T}\v{w}_{m}^{h}=0$ for every vector $\v{v}_{m}^{f}$ in $V_{m}$, which means
\begin{equation}
\begin{aligned}
\sum_{f=1}^{r}(\v{v}_{m}^{f})^{T}\v{w}_{m}^{h}\v{v}_{m}^{f}=\v0.
\end{aligned}
\label{eq:rr_loss_11}
\end{equation}
As a consequence, the constraint \myrefeq{eq:rr_loss_1} is only partially fulfilled:
\begin{equation}
\begin{aligned}
\v{c}_{m_h}(\sum_{f=1}^{r}(\v{v}_{m}^{f})^{T}\v{w}_{m}^{h}\v{v}_{m}^{f}-\v{w}_{m}^{h})=
-\v{c}_{m_h}\v{w}_{m}^{h} \ .
\end{aligned}
\label{eq:rr_loss_8}
\end{equation}
\nilsE{As the $\v{w}_{m}^{h}$ have unit length,
the} factors $\v{c}_{m}$ determine the contribution of a datum to the overall loss.
A feature $\v{w}_{m}^{h}$ that appears multiple times in the input 
data will have a correspondingly larger factor in $\v{c}_{m}$ and hence
will more strongly contribute to $\Loss{\text{RR}}$.
The $L^2$ formulation of \myrefeq{eq:rr_loss_o} leads to 
\nilsE{the largest contributors being minimized most strongly,
and hence } the repeating features of the data, i.e.,
dominant features, need to be represented in $V_{m}$ to minimize the racecar loss.

In summary, to  minimize $\Loss{\text{RR}}$, $V_{m}$ is driven towards containing $r$ orthogonal 
vectors $\v w_{m}^{h}$ which represent the most frequent features of the input data, i.e. the dominant features.
It is worth emphasizing that above $\mathcal{B}_{m}$ is only an auxiliary basis, i.e., the derivation does not depend on any particular choice of $\mathcal{B}_{m}$.

\subsection{Examples of Network Architectures with Racecar Training}
\label{sec:structure}
To specify NN architectures, we use the following notation: 
$C(k,l,q)$, and $D(k,l,q)$ denote convolutional and deconvolutional operations, respectively, while fully connected layers are denoted with $F(l)$, where $k$, $l$, $q$ denote kernel size, output channels and stride size, respectively. The bias of a CNN layer is denoted with $b$. $I/O(z)$ denote $input/output$, their dimensionality is given by $z$. $I_{r}$ denotes the input of the reverse pass network. $tanh$, $relu$, $lrelu$ denote hyperbolic tangent, ReLU, and leaky ReLU activation functions (AF), where we typically use a leaky tangent of 0.2 for the negative half-space. $UP$, $MP$ and $BN$ denote $2\times$ nearest-neighbor up-sampling, max pooling with $2\times2$ filters and stride 2, and batch normalization, respectively. 

Below we provide additional examples how to realize the racecar loss $\Loss{\text{racecar}}$ in a 
neural network architecture. As explained in the main document, 
the constraint \myrefeq{eq:rr_loss_o} is formulated via 
\begin{equation}
\Loss{\text{racecar}}=\sum_{m=1}^{n}{\lambda_{m}\left \| \datum{m} - \datumt{m}{'}\right \|}_F^2 , 
\label{eq:racecarloss}
\end{equation}
with $\datum{m}$, and $\lambda_{m}$ denoting the vector of 
\nilsE{activated intermediate data in layer $m$ from the forward pass, and a scaling factor, respectively. 
$\datumt{m}{'}$ denotes the activations of layer $m$ from the reverse pass.
E.g., let $L_m()$ denote the operations of a layer $m$ in the foward pass, and $L_m'()$ the corresponding 
operations for the reverse pass. Then $\datumt{m+1}{}=L_m(\datumt{m}{})$, 
and $\datumt{m}{'}=L_{m}'(\datumt{m+1}{'})$. 
}

When \myrefeq{eq:racecarloss} is minimized, we obtain activated intermediate content 
during the reverse pass that reconstructs the values computed in the forward pass, i.e. 
$\datumt{m+1}{'}=\datum{m+1}$ holds.
Then $\datumt{m}{'}$ can be reconstructed from the incoming activations from the reverse pass, i.e., $\datumt{m+1}{'}$,
or from the output of layer $m$, i.e., $\datum{m+1}$. 
Using $\datumt{m+1}{'}$ results in a global coupling of input and output
throughout all layers, i.e., the {\em full racecar loss} variant. On the other hand, 
$\datum{m+1}$ yields a variant that ensures local reversibility of each layer,
and yields a very similar performance, as we will demonstrate below. We employ this 
{\em layer-wise racecar loss} for networks without a 
unique, i.e., bijective, connection
between two layers. Intuitively, when inputs cannot be reliably reconstructed from outputs.

\begin{figure}[bt!]
\vspace{-2mm}
	\begin{center}
		\includegraphics[width=0.7\linewidth]{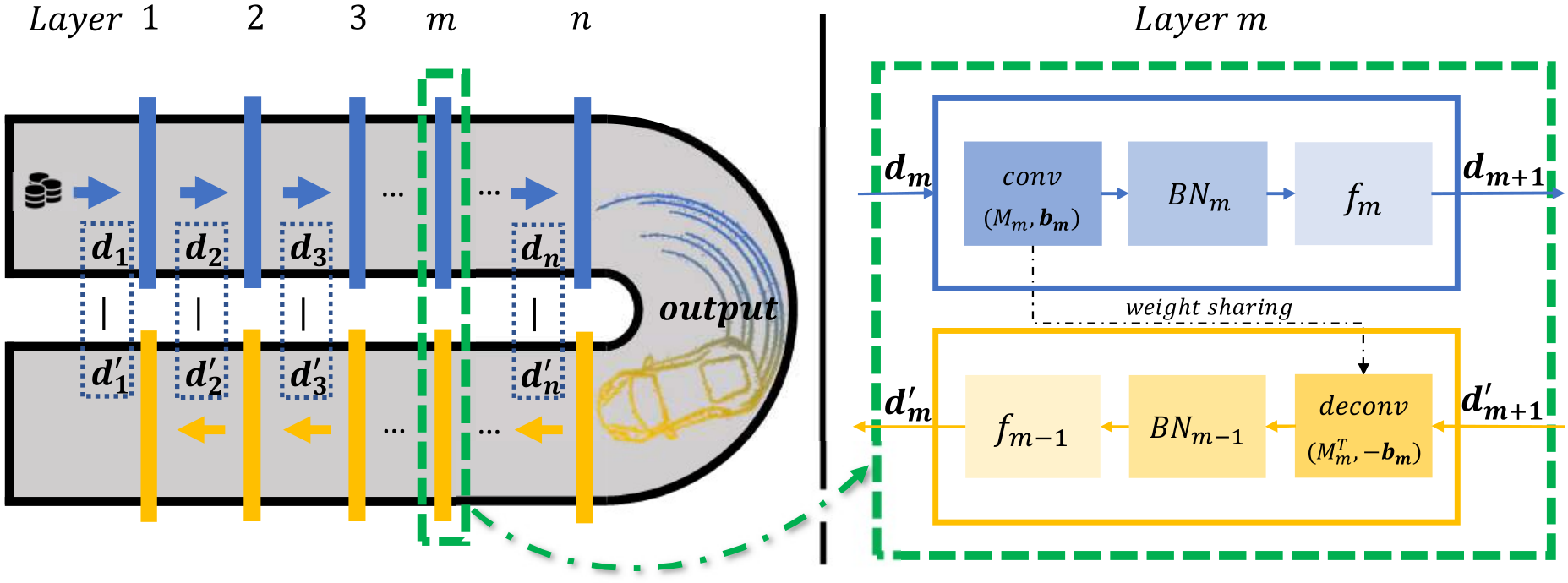}
	\end{center}
	\vspace{-4mm}
	\caption{\footnotesize{
	Left: An overview of the regular forward pass (blue) and the corresponding 
   reverse pass (yellow). The right side illustrates how  
	parameters are reused for a convolutional layer.
	$conv/deconv$ denote convolution/deconvolutional operations.
	$f_{m}$ and $BN_{m}$ denote the activation function and batch normalization of layer $m$, respectively}.
	Shared kernel and bias are represented by ${M}_{m}$ and $\v{b}_{m}$.
	} 
	\label{fig:pipeline}
	\vspace{-2mm}
\end{figure}

\paragraph{Full Racecar Training:} 
An illustration of a CNN structure with AF and BN and a full racecar loss is shown in \myreffig{fig:pipeline}. 
While the construction of the reverse pass is straight-forward for all standard operations,
i.e., fully connected layers, convolutions, pooling, etc., 
slight adjustments are necessary for AF and BN.
It is crucial for our formulation that $\datum{m}$ and $\datumt{m}{'}$ 
contain the same latent space content in terms of range and dimensionality,
such that they can be compared in the loss.
Hence, we use the BN parameters and the AF of layer $m-1$ from the forward pass 
for layer $m$ in the reverse pass. An example is shown in \myreffig{fig:more_blob_SVD}. 

To illustrate this setup, we consider an example network employing convolutions with mixed 
AFs, BN, and MP. Let the network receives a field of $32^2$ scalar values as input. 
From this input, 20, 40, and 60 feature maps are extracted in the first three layers.
Besides, the kernel sizes are decreased from $5 \times 5$ to $3 \times 3$.
To clarify the structure, we use ReLU activation for the first convolution, 
while the second one uses a hyperbolic tangent, and the third one a sigmoid function. 
With the notation outlined above, the first three layers of the network are  
\begin{equation}
 \resizebox{0.6\hsize}{!}{$
 \begin{aligned}
I(32,32,1) \nilsE{=} \ & \datum{1} \rightarrow C_{1}(5,20,1)+\v{b}_1\rightarrow BN_{1} \rightarrow \text{relu} \\
 \rightarrow \ & \datum{2}\rightarrow MP \rightarrow C_{2}(4,40,1)+\v{b}_2\rightarrow BN_{2} \rightarrow \text{tanh} \\
 \rightarrow \ & \datum{3}\rightarrow MP \rightarrow C_{3}(3,60,1)+\v{b}_3\rightarrow BN_{3} \rightarrow \text{sigm}  \\
 \rightarrow \ & \datum{4}\rightarrow ...
 \end{aligned}
 $}
\end{equation}

The reverse pass for evaluating the racecar loss re-uses all weights of the 
forward pass and ensures that all intermediate 
vectors of activations, $\datum{m}$ and $\datumt{m}{'}$, have the same size 
and content in terms of normalization and non-linearity. We always consider 
states after activation for $\Loss{\text{racecar}}$. Thus, $\datum{m}$ denotes 
activations before pooling in the forward pass and $\datumt{m}{'}$ contains
data after up-sampling in the reverse pass, in order to ensure matching dimensionality.
Thus, the last three layers of the reverse network for computing $\Loss{\text{racecar}}$ take the form:
\begin{equation}
 \resizebox{0.5\hsize}{!}{$
 \begin{aligned}
  ... & \rightarrow \datumt{4}{'} \rightarrow -\v{b}_3\rightarrow D_{3}(3,40,1) \rightarrow BN_{2} \rightarrow \text{tanh} \rightarrow UP \\
 & \rightarrow \datumt{3}{'}\rightarrow -\v{b}_2\rightarrow D_{2}(4,20,1) \rightarrow BN_{1} \rightarrow \text{relu} \rightarrow UP  \\
 & \rightarrow \datumt{2}{'}\rightarrow -\v{b}_1\rightarrow D_{1}(5,3,1) \\ 
  & \rightarrow \datumt{1}{'} = O(32,32,1).
 \end{aligned}
 $}
\end{equation}
Here, the de-convolutions $D_{x}$ in the reverse network share weights with $C_{x}$ in the forward network.
I.e., the $4 \times 4 \times 20 \times 40$ weight matrix of $C_2$ is reused in its transposed form as a 
$4 \times 4 \times 40 \times 20$ matrix in $D_2$.
Additionally, it becomes apparent that AF and BN of layer 3 from the forward 
pass do not appear in the listing of the three last layers of the reverse pass. This is caused by the fact 
that both are required to establish the latent space of the fourth layer. Instead, $\datum{3}$ in our example represents 
the activations after the second layer (with $BN_{2}$ and $tanh$), and hence the reverse pass for $\datumt{3}{'}$
reuses both functions.
This ensures that $\datum{m}$ and $\datumt{m}{'}$ contain the same latent space content in terms of range and dimensionality, and can be compared in \myrefeq{eq:racecarloss}. 

For the reverse pass, we additionally found it beneficial to employ an AF for the very last layer if the output space has suitable content. E.g., 
for inputs in the form of RGB data we employ an additional activation 
with a ReLU function for the output to ensure the network generates
only positive values. The architectures in Tab. 11, 13, 15, 19, 22, and 24 use such an activation.

\paragraph{Layer-wise Racecar Training:}
In the example above, we use a full racecar training with $\datumt{m+1}{'}$ to reconstruct the activations $\datumt{m}{'}$. 
The full racecar structure establishes a slightly stronger relationship among the racecar loss terms of different layers, 
and allows earlier layers to decrease the accumulated loss of later layers.
However, if the architecture of the original network makes use of operations between layers that are not bijective,
we instead use the layer-wise racecar loss. E.g., this happens for residual connections with an addition 
or non-invertible pooling operations such as max-pooling. In the former, we cannot uniquely determine 
the $b,c$ in $a=b+c$ given only $a$. And unless special care is taken \cite{bruna2013signal}, 
the source neuron of an output is not known for regular max-pooling operations.
Note that our racecar loss has no problems with 
irreversible operations within a layer, e.g., most 
convolutional or fully-connected layers typically are not fully invertible. In all these cases the loss 
will drive the network towards a state that is as-invertible-as-possible for the 
given input data set. However, this requires 
a reliable vector of target activations in order to apply the constraints. If the {\em connection} betweeen layers
is not bijective, we \nilsE{cannot reconstruct} this target for the constraints, as in the examples given above.

In such cases, we regard every layer as an individual unit to which we apply the racecar loss
by building a layer-wise reverse pass.
For example, given a simple convolutional architecture with 
\begin{equation}
\datum{1} \rightarrow C_{1}(5,20,1)+\v{b}_1 = \datum{2} 
\label{eq:forward_1}
\end{equation}
 in the forward pass, we calculate $\datum{1}^{'}$ with 
 \begin{equation}
(\datum{2}-\v{b}_1)\rightarrow D_{1}(5,3,1)=\datum{1}^{'},
\label{eq:forward_2}
\end{equation}
We, e.g., use this layer-wise loss in the Resnet110 network shown in Tab. 17.

\begin{figure*}[t!]
    \begin{center}
		\includegraphics[width=0.99\linewidth]{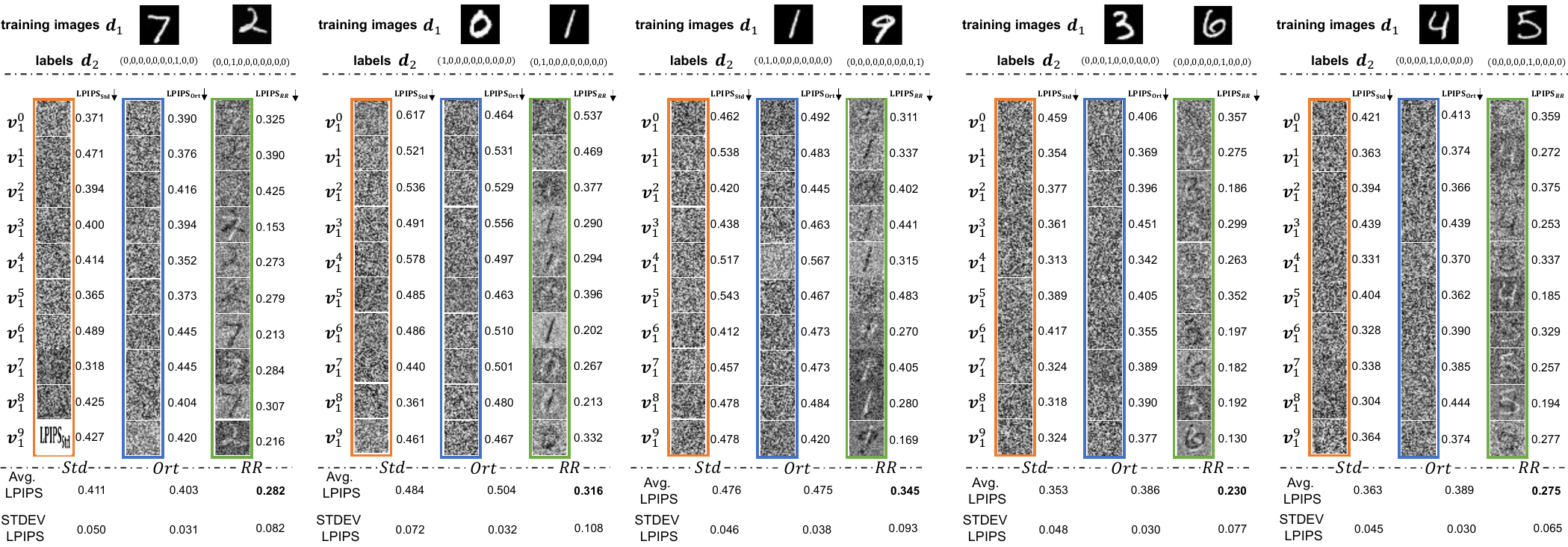}
	\end{center}
	\caption{\footnotesize{SVD of the $M_1$ matrix for five tests with random two digit images as training data. LPIPS distances \cite{zhang2018unreasonable} of \vrrPow{}{} are consistently lower than \vstd{} and \vort{}.}}
	\label{fig:more_MNIST_SVD}
\end{figure*}
\begin{figure*}[t!]
\vspace{-3mm}
    \begin{center}
		\includegraphics[width=0.99\linewidth]{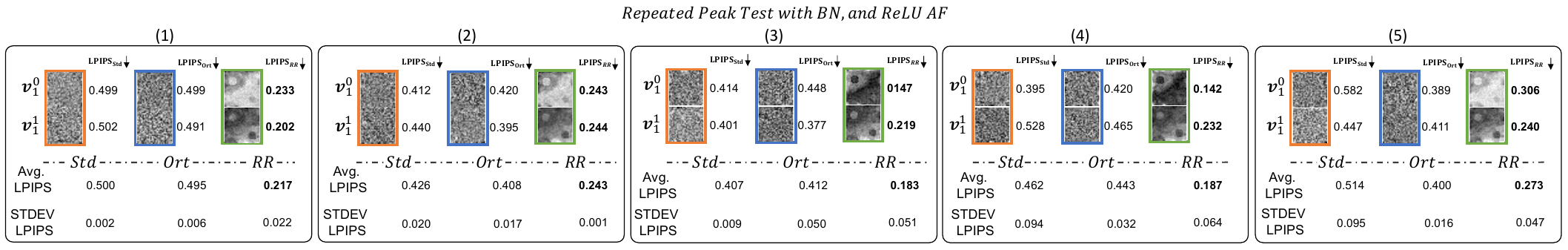}
	\end{center}
	\caption{\footnotesize{Five repeated tests with the peak data shown in Sec. 3 of the main paper. \vrrPow{A}{} robustly extracts dominant features from the data set. The two singular vectors strongly resemble the two peak modes of the training data. This is confirmed by the LPIPS measurements.}}
	\label{fig:repeated_blob_SVD}
\end{figure*}
\subsection{MNIST and Peak tests}
\label{sec:Blobtestapp}
\begin{wrapfigure}{hR}{0.65\linewidth}
\vspace{-7mm}
    \begin{center}
		\includegraphics[width=0.99\linewidth]{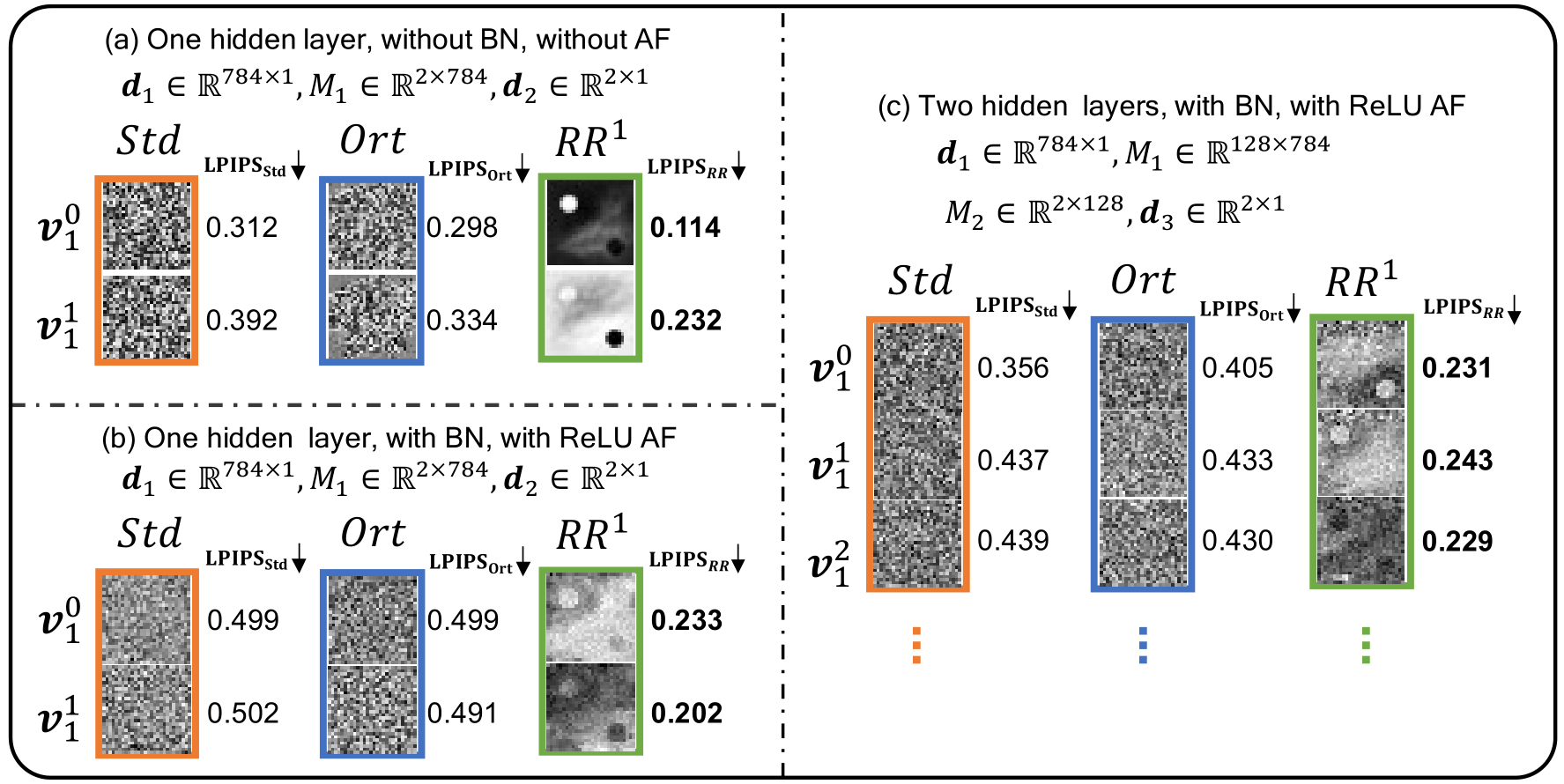}
	\end{center}
	\vspace{-3mm}
	\caption{\footnotesize{Right singular vectors of $M_{1}$ for 
    peak tests with different network architectures: from a single linear layer to two layers with BN and AF.
	Across the three architectures, \vrrPow{A}{} successfully extracts dominant 
    features. In contrast to \vstd{} and \vort{}, the singular vectors contain salient structures.
    }}
	\label{fig:more_blob_SVD}
\end{wrapfigure}
Below we give details for the MNIST and {\em peak} tests from Sec. 3 of the main paper. 
\paragraph{MNIST Test}:
For the MNIST test, additional SVDs of the weight matrices of trained models can be seen in \you{\myreffig{fig:teaser} and} \myreffig{fig:more_MNIST_SVD}. The LPIPS scores (lower being better) show that features embedded in the weights of \vrrPow{}{} are significantly closer to the training data set than \vstd{} and \vort{}. All MNIST models are trained for 1000 epochs with a learning rate of $0.0001$, and $\lambda=1e-5$ for \vrrPow{A}{}.
The NN architecture of the tests in this section is given in Tab. 8.

\begin{wrapfigure}{hR}{0.6\linewidth}
\vspace{-4mm}
    \begin{center}
		\includegraphics[width=0.99\linewidth]{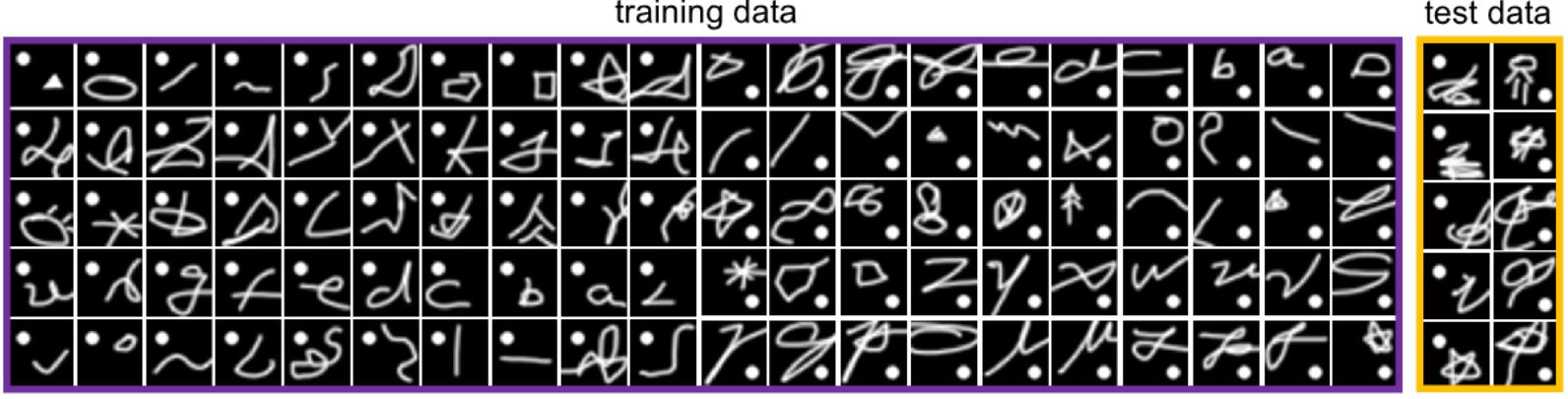}
	\end{center}
	\caption{\footnotesize{Data set used for the {\em peak} tests.}}
	\label{fig:blob_data}
\end{wrapfigure}

\paragraph{{\em Peak} Test}:
For the {\em Peak} test we generated a data set of 110 images shown in \myreffig{fig:blob_data}. 55 images contain a peak located in the upper left corner of the image. The other 55 contain a peak located in the bottom right corner. We added random scribbles in the images to complicate the task. All 110 images were labeled with a one-hot encoding of the two possible positions of the peak.
We use 100 images as training data set, and the remaining 10 for testing. All peak models are trained for 5000 epochs with a learning rate of $0.0001$, with $\lambda=1e-6$ for \vrrPow{A}{}.
\nilsE{To draw reliable conclusions, we show results for five repeated runs here.} 
The neural network in this case contains one fully connected layer, with BN and ReLU activation. The results are shown in \myreffig{fig:repeated_blob_SVD}, with both peak modes being consistently
embedded into the weight matrix of \vrrPow{A}{}, while regular and orthogonal training show primarily random singular vectors.

We also use different network architectures in \myreffig{fig:more_blob_SVD}
to verify that the \nilsE{dominant features} are successfully extracted when using more complex network structures.
Even for two layers with BN and ReLU activations, the racecar training
clearly extracts \nilsE{the two modes} of the training data. The visual resemblance is slightly 
reduced in this case, as the network has the freedom to embed the features in both layers.
Across all three cases (for which we performed 5 runs each), the racecar training clearly 
outperforms regular training and the orthogonality constraint in terms of extracting and 
embedding the dominant structures of the training data set in the weight matrix.

Overall, our experiments confirm the theory behind the racecar formulation.
They additionally show that racecar training in combination with a subsequent SVD of the network weights
yields a simple and convenient method to give humans intuition about the features learned by 
a network.

\section{Evaluation via Mutual Information}
\label{sec:mutualinformation}

This section gives details of the mutual information and disentangled representation tests from Sec. 4 of the main paper.

\subsection{Mutual Information test}
\label{sec:MItestapp}

We now evaluate our approach in terms of mutual information (MI), which
measures the dependence of two random variables, i.e., 
higher MI means that there is more shared information between two parameters. 
More formally, the mutual information $I(X;Y)$ of
random variables $X$ and $Y$ measures how different the joint distribution of $X$ and $Y$ is w.r.t. the product of their marginal distributions, i.e., the Kullback-Leibler divergence 
$I(X;Y) = \text{KL}[P_{(X,Y)}||P_X P_Y]$, where $\text{KL}$ denotes the Kullback-Leibler divergence. 
Let $I(X;\data{m})$ denote the mutual information between the 
activations of a layer $\data{m}$ and input X. 
Similarly $I(\data{m};Y)$ denotes the MI between layer $m$ and the output $Y$. We use {\em MI planes} in the main paper, which show $I(X;\data{m})$ and $I(\data{m};Y)$ in a 2D graph for the activations of each layer $\data{m}$ of a network after training. This visualizes how much information about input and output distribution
is retained at each layer, and how these relationships change within the network. 
For regular training, the information bottleneck principle \cite{tishby2015deep} states that early
layers contain more information about the input, i.e.,
show high values for $I(X;\data{m})$ and $I(\data{m};Y)$. 
Hence in the MI plane visualizations, these layers are often visible at the top-right corner. 
Later layers typically share a large amount of information with the output
after training, i.e. show large $I(\data{m};Y)$ values, and
correlate less with the input (low $I(X;\data{m})$).
Thus, they typically show up in the top-left corner of the MI plane graphs. 

\paragraph{Training Details:} 
We use the same numerical studies as in \cite{shwartz2017opening} as task $A$,
i.e. a regular feed-forward neural network with 6 fully-connected layers. The 
input variable $X$ contains 12 binary digits that represent 12 uniformly distributed points 
on a 2D sphere. The learning objective is to discover binary decision rules which are 
invariant under $O(3)$ rotations of the sphere. $X$ has 4096 different patterns, which 
are divided into 64 disjoint orbits of the rotation group, forming a minimal sufficient 
partition for spherically symmetric rules \cite{kazhdan2003rotation}. 
To generate the input-output distribution $P(X,Y)$, We apply the stochastic rule 
$p(y=1|x)=\Psi (f(x)-\theta), (x\in X, y\in Y$), where $\Psi$ is a standard sigmoidal 
function $\Psi(u)=1/(1+exp(-\gamma u))$, following  \cite{shwartz2017opening}. We then use 
a spherically symmetric real valued function of the pattern $f(x)$, evaluated through its 
spherical harmonics power spectrum \cite{kazhdan2003rotation}, and compare with a threshold 
$\theta$, which was selected to make $p(y=1)=\sum_{x}p(y=1|x)p(x)\approx 0.5$, with uniform 
$p(x)$. $\gamma$ is high enough to keep the mutual information $I(X;Y) \approx  0.99$ bits. 

For the transfer learning task $B$, we reverse output labels to check whether the model learned specific or generalizing features. E.g., if the output is [0,1] in the original data set, we swap the entries to [1,0].
80\% of the data (3277 data pairs) are used for training and rests (819 data pairs) are used for testing. The forward and reverse pass architectures of the fully connected neural networks are given in Tab. 9. Hyperparameters used for training are listed in Tab. 10. Models are trained using cross-entropy as the base loss function. 

For the MI comparison in Fig. 4 of the main paper, we discuss models before and after fine-tuning separately, in order to illustrate the effects of regularization. We include a regular model \vstd{A}, one with orthogonality constraints \vort{A}, and our regular racecar model \vrrPow{A}{}, all before fine-tuning.
For the regular racecar model \vrrPow{A}{} all layers are constrained to be recovered in the backward pass. 
We additionally include the version {\color{rr1Col}$\text{RR}_\text{A}^{1}$}, i.e. a model trained with only one racecar loss term ${\lambda_{1}| \datum{1} - \datumt{1}{'}|}_2$, which means that only the input is constrained to be recovered. 
Thus, {\color{rr1Col}$\text{RR}_\text{A}^{1}$} represents a simplified version of our approach which receives no constraints that intermediate results of the forward and backward pass should match. 
For \vort{A}, we used the  Spectral Restricted Isometry Property (SRIP) regularization \cite{bansal2018can},
\begin{equation}
    \begin{aligned}
        \mathcal{L}_{\text{SRIP}} = \beta \sigma (W^{T}W-I),
    \end{aligned}
    \label{eq:ortho}
\end{equation}
where $W$ is the kernel, $I$ denotes an identity matrix, and $\beta$ represents the regularization coefficient. 
$\sigma (W)=sup_{z\in \mathbb{R}^{n},z\neq 0}\frac{\left \|W_{z}  \right \|}{\left \|z  \right \|}$ denotes the spectral norm of $W$. 

As explained in the main text, all layers of the first stage, i.e. from \vrrPow{A}{}, {\color{rr1Col}$\text{RR}_\text{A}^{1}$}, \vort{A} and \vstd{A} are reused for training the fine-tuned models without regularization, i.e. \vrrPow{AA}{}, {\color{rr1Col}$\text{RR}_\text{AA}^{1}$}, \vort{AA} and \vstd{AA}. 
Likewise, all layers of the transfer task models \vrrPow{AB}{}, {\color{rr1Col}$\text{RR}_\text{AB}^{1}$}, \vort{AB} and \vstd{AB} are initialized from the models of the first training stage.

\paragraph{Analysis of Results:} 
We first compare the version only constraining input and output reconstruction ({\color{rr1Col}$\text{RR}_\text{A}^{1}$})
and the full racecar loss version \vrrPow{A}{}.
Fig. 4(b) of the main paper shows that all points of \vrrPow{A}{} are located in a central region of the MI place, which means that all layers successfully encode information about the inputs as well as the outputs. This also indicates that every layer contains a similar amount of information about $X$ and $Y$,
and that the path from input to output is similar to the path from output to input.
The points of {\color{rr1Col}$\text{RR}_\text{A}^{1}$}, on the other hand, form a diagonal line. 
I.e., this network has different amounts of mutual information across its layers, and potentially a very different path for each direction. 
This difference in behavior is caused by the difference of the constraints in these two versions: 
{\color{rr1Col}$\text{RR}_\text{A}^{1}$} is only constrained to be able to regenerate its input, while the full racecar
loss for \vrrPow{A}{} ensures that the network learns features which are beneficial for both directions.
This test highlights the importance of the constraints throughout the depth of a network in our formulation.
In contrast, the $I(X;\data{})$ values of later layers for \vstd{A} and \vort{A} exhibit small values (points near the left side), while $I(\data{};Y)$ is high throughout.
This indicates that the outputs were successfully encoded and that increasing amounts of information about the inputs are discarded. Hence, more specific features about the given output data-set are learned by \vstd{A} and \vort{A}. 
This shows that both models are highly specialized for the given task, and potentially perform worse when applied to new tasks.

During the fine-tuning phase for task $A$ (i.e. regularizers being disabled), all models focus on the output and maximize $I(\data{};Y)$. There are differences in the distributions of the points along the y-axis, i.e., how much MI with the output is retained, as shown in Fig. 4(c) of the main paper.
For model \vrrPow{AA}{}, the $I(\data{};Y)$ value is higher than for \vstd{AA}, \vort{AA} and {\color{rr1Col}$\text{RR}_\text{AA}^{1}$}, which means outputs of \vrrPow{AA}{} are more closely related to the outputs, i.e., the ground truth labels for task $A$. Thus, \vrrPow{AA}{} outperforms the other variants for the original task.

In the fine-tuning phase for task $B$, \vstd{AB} stands out with very low accuracy in Fig. 5 of the main paper. This model from a regular training run has large difficulties to adapt to the new task. Model \vort{AB} also performs worse than \vstd{B}. \vrrPow{AB}{} shows the best performance in this setting, demonstrating that our loss formulation helped to learn more generic features from the input data. This improves the performance for related tasks such as the inverted outputs used for $B$.

We also analyze the two variants of racecar training training, the layer-wise variant {\color{lrrCol}$\text{lRR}_\text{A}$} and the full version \vrrPow{A}{} in terms of mutual information.
\myreffig{fig:informationplane} shows the MI planes for these two models, also showing \vrrPow{A}{1} for comparison. 
Despite the local nature of  {\color{lrrCol}$\text{lRR}_\text{A}$} it manages to establish MI for the majority of
the layers, as indicated by the cluster of layers in the center of the MI plane. 
Only the first layer moves 
towards the top right corner, and the second layer is affected slightly. I.e., these layers exhibit a stronger relationship with the distribution of the outputs. Despite this, the overall performance when fine-tuning or for the task transfer
remains largely unaffected, e.g., the {\color{lrrCol}$\text{lRR}_\text{A}$} still clearly outperforms \vrrPow{A}{1}.
This confirms our choice to use the full racecar training when network connectivity permits,
and employ the layer-wise version in all other cases.

Numerical accuracies for all models discussed in this section are listed in Tab. 1 and Tab. 2.
Mutual information values, i.e., $I(X,\data{m})$ and $I(\data{m},Y)$, for all models and layers are provided in Tab. 26.

\begin{figure*}
    \vspace{-4mm}
    \centering
        \begin{overpic}[width=0.99\linewidth]{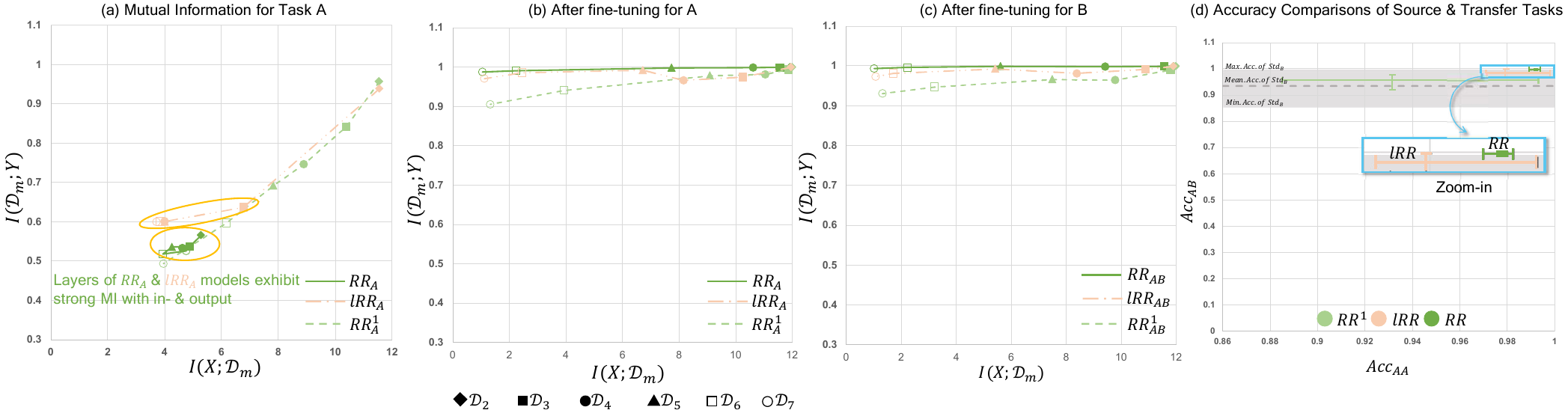}
        \end{overpic}
    \caption{\footnotesize{
    (a-c) MI plane comparisons for layer-wise 
    ({\color{lrrCol}$\text{lRR}_\text{A}$}) versus full models (\vrrPow{A}{}). 
    Points on each line correspond to layers of one type of model.
    a) MI Plane for task A. 
    All points of \vrrPow{A}{} and the majority of points for {\color{lrrCol}$\text{lRR}_\text{A}$} (five out seven) 
    are located in the center of the graph, i.e., successfully connect in- and ouput distributions. 
    b,c): After fine-tuning for A/B. 
    The last layer $\data{7}$ of \vrrPow{AA}{} builds the strongest relationship with $Y$. $I(\data{7};Y)$ of {\color{lrrCol}$\text{lRR}_\text{A}$} is only slightly lower than \vrrPow{AA}{}.
    d): Accuracy comparisons among different models: \vrrPow{AA}{} yields the highest performance, while {\color{lrrCol}$\text{lRR}_\text{A}$} performs similarly with \vrrPow{AA}{}.
    } } \label{fig:informationplane}
\end{figure*} 

\subsection{Disentangled Representations}
\label{sec:infoGANtestapp}
The InfoGAN approach \cite{chen2016infogan} demonstrated the possibility to control the output of generative models via maximizing mutual information between outputs and structured latent variables. However, mutual information is very hard to estimate in practice  \cite{walters2009estimation}. 
The previous section and Fig. 4(b) of the main paper demonstrated that models from racecar training (both {\color{rr1Col}$\text{RR}_\text{A}^{1}$} and \vrrPow{A}{}) can increase the mutual information between network inputs and outputs. Intuitively, racecar training explicitly constrains the model to reocver an input given an output, which directly translates into an increase of mutual information between input and output distributions compared to regular training runs.
For highlighting how racecar training can yield disentangled representations (as discussed 
in the later paragraphs of Sec. 4 of the main text), we follow the experimental setup of 
InfoGAN \cite{chen2016infogan}: the input dimension of our network is 74, containing 1 ten-dimensional category code $c_{1}$, 2 continuous latent codes $c_{2},c_{3} \sim \mathcal{U}(-1,1)$ and 62 noise variables. 
Here, $\mathcal{U}$ denotes a uniform distribution.

\begin{figure*}[t!]
    \begin{center}
		\includegraphics[width=0.99\linewidth]{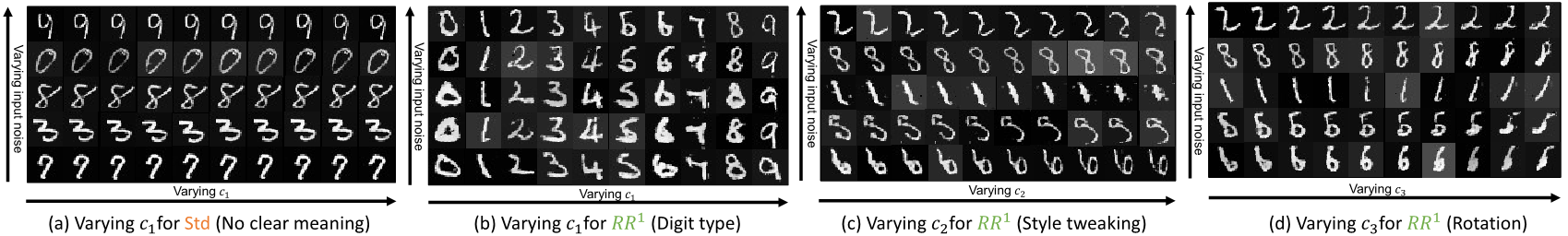}
	\end{center}
     \caption{\footnotesize{
     Additional results for the disentangled representations with the MNIST data: 
     For every row in the figures, we vary the corresponding latent code (left to right), while keeping 
     all other inputs constant. Different rows indicate a different random noise input. For example, in
     (b): every column contains five results which are generated with different noise samples, but the same latent codes $c_{1\sim3}$. In every row, 10 results are generated with 10 different values of $c_{1}$, which correspond to one digit each for (b). 
     (a): For a regular training (\vstd{}), no clear correspondence between $c_{1}$ and the outputs are apparent (similarly for $c_{2,3}$).
     (c): Different $c_{2}$ values result in a tweaked style, while $c_{3}$ controls the orientation of the digit, as shown in (d).
     Thus, in contrast to \vstd{}, the racecar model learns a meaningful, disentangled representation.
     } } \label{fig:infoGAN}
\end{figure*}

\paragraph{Training Details:} 
As InfoGAN focuses on structuring latent variables 
and thus only increases the mutual information between latent variables and the output, we also 
focus the racecar training on the corresponding latent variables. 
I.e., the goal is to maximize their mutual information with the output of the generative model.
Hence, we train a model {\color{rr1Col}$\text{RR}_\text{}^{1}$} for which 
only latent dimensions $c_{1},c_{2},c_{3}$ of the input layer are involved in 
the racecar loss. We still employ a full reverse pass structure in the neural network architecture.
$c_{1}$ is a ten-dimensional category code, which is used for controlling the output digit category, 
while $c_{2}$ and $c_{3}$ are continuous latent codes, to represent (previously unknown) key
properties of the digits, such as orientation or thickness. 
Building relationship between $c_{1}$ and outputs is more difficult than for $c_{2}$ or $c_{3}$, since the 
10 different digit outputs need to be encoded in a sinlge continuous variable $c_{1}$.
Thus, for the corresponding racecar loss term for $c_{1}$ we use a slightly larger $\lambda$ factor (by $33\%$) than for $c_{2}$ and $c_{3}$.
The forward and reverse pass architectures of the network are given in Tab. 11, and hyperparameters are listed in Tab. 12. Details of our results are shown in \myreffig{fig:infoGAN}. Models are trained using a GAN loss \cite{goodfellow2014generative} as the loss function for the outputs.

\paragraph{Analysis of Results}:
In \myreffig{fig:infoGAN} we show additional results for the disentangling test case.
It is visible that the racecar training with the {\color{rr1Col}$\text{RR}_\text{}^{1}$} model
yield distinct and meaningful latent space dimensions for $c_{1,2,3}$. While
$c_{1}$ controls the digit, $c_{2,3}$ control the style and orientation of the digits.
For comparison, a regular training run with model \vstd{} does result in meaningful or visible
changes when adjusting the latent space dimensions. This illustrates how strongly racecar training can 
shape the latent space, and in addition to an intuitive embedding of dominant features, 
yield a disentangled representation.

\section{Additional Experiments}
\label{sec:addexperiment}

In this section we present several additional experiments that were not included in the main document
due to space constraints.

\subsection{MNIST Classification}

\label{sec:MNISTtestapp}
\begin{wrapfigure}{hR}{0.45\linewidth}
\vspace{-4mm}
    \centering
        \includegraphics[width=0.99\linewidth]{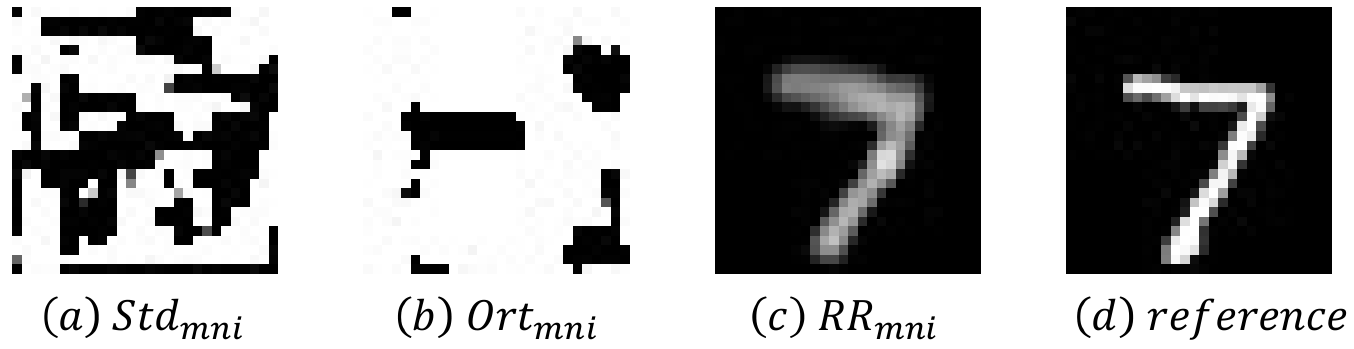}
    \caption{\footnotesize{Comparisons between reconstructed inputs. 
    Only \vrrPow{$mni$}{} recovers most of the input information successfully.}
    } \label{fig:MNISTreinput}
\end{wrapfigure}

The MNIST data set is a commonly used data set for hand written digit classification \cite{lecun1998gradient}. 
We train three types of models with the regular MNIST data set: 
one with racecar loss \vrrPow{}{}, a regular model \vstd{}, and one with orthogonal constraints \vort{} \cite{bansal2018can} for comparison.
Usually, convolutional layers take up most of the model parameters, so we correspondingly compute the racecar loss for the convolutional layers, omitting the fully connected layers that would be required for class label inference.
We follow the common practice to first train models with regularization and then fine-tune without \cite{bansal2018can}. As we discuss model states before fine-tuning in this section, we use the following naming scheme with suffixes {\em mni, MNI} and {\em n-MNI} to distinguish the different phases.
E.g., with racecar training:
\vrrPow{$mni$}{} for training runs on MNIST with regularization,
\vrrPow{MNI}{} for fine-tuning on MNIST without regularization using
\vrrPow{$mni$}{} as starting point,
and \vrrPow{n-MNI}{} for fine-tuning on a transfer task without regularization, likewise using
\vrrPow{$mni$}{} as starting point.

\begin{wrapfigure}{hR}{0.5\linewidth}
 \vspace{-5mm}
    \begin{center}
		\includegraphics[width=0.99\linewidth]{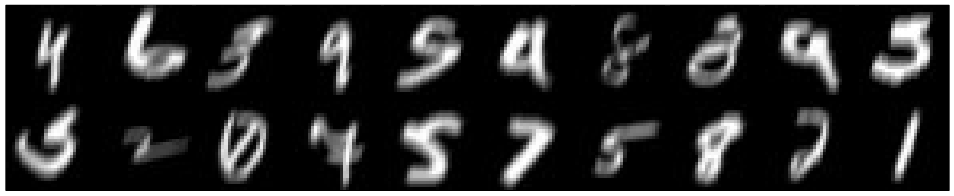}
	\end{center}
	\caption{\footnotesize{Example data from n-MNIST.}}
	\label{fig:MNdata}
	 \vspace{-2mm}
\end{wrapfigure}

\paragraph{Training Details:} The regular MNIST data set \cite{lecun1998gradient} contains $55k$ images for training and $10k$ images for testing. For the n-MNIST motion blur data set \cite{basu2017learning}, there are $60k$ images for training and $10k$ images for testing. All images have a size of $28\times28$.
Example data from n-MNIST that illustrates the motion blur is shown in \myreffig{fig:MNdata}.
Details of the network architectures are shown in Tab. 15, with hyperparameters given in Tab. 16. 
The models are trained using cross-entropy as base loss function,
and all three convolutional layers are used for the racecar loss.

\paragraph{Analysis of Results}: 
To highlight the properties of our algorithm, we show comparisons 
between the original input $\datum{1}$ and the reconstructed inputs $\datumt{1}{'}$ in \myreffig{fig:MNISTreinput}
for the models \vrrPow{$mni$}{}, \vstd{$mni$}, and \vort{$mni$}.
These models were trained with an enabelde regularization, i.e., before fine tuning.
For racecar training, most of the features from the input are recovered, while
trying to invert the network in the same way for a
regular training run or a training with orthogonal constraints largely fails.
In both cases, features are extracted only according to the digit classification task,
and both models discard information unrelated to this goal.

\begin{wrapfigure}{hR}{0.45\linewidth}
\vspace{-4mm}
    \centering
    \includegraphics[width=0.99\linewidth]{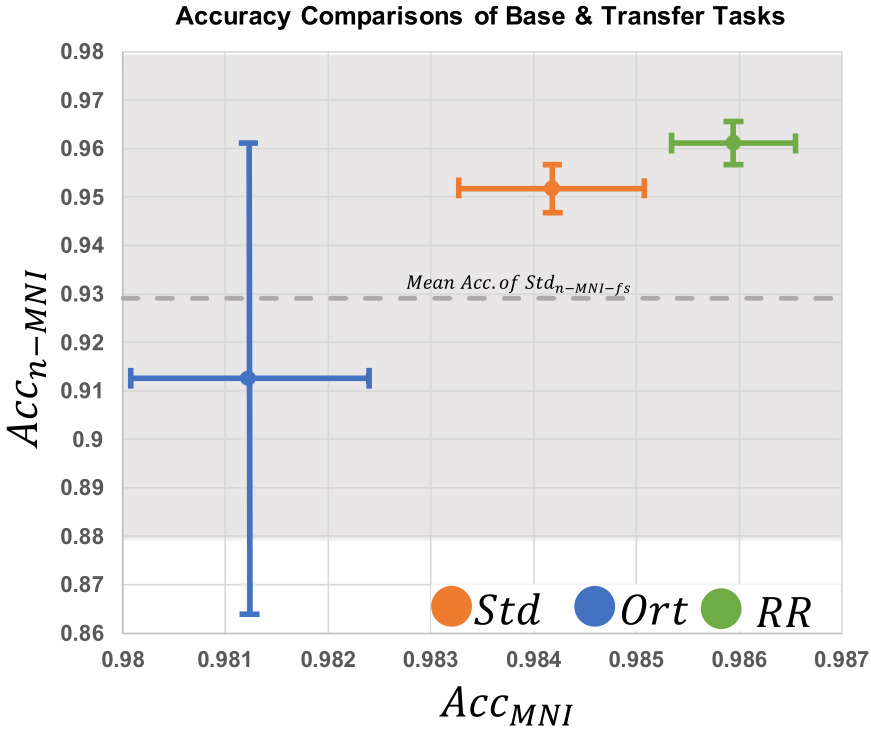}
    \vspace{-3mm}
    \caption{\footnotesize{Accuracy comparisons for original task (Acc$_{\text{MNI}}$) and transfer learning task (Acc$_{\text{n-MNI}}$). The dashed grey line and grey region represent the accuracy of 
    baseline models, \vstd{n-MNI-fs} trained from scratch for n-MNIST. \vrrPow{MNI}{} and \vrrPow{n-MNI}{} outperform this baseline, and achieve the best performance for the source task and transfer learning task, respectively.}
    } \label{fig:MNacc}
\vspace{-4mm}
\end{wrapfigure}

For the final MNIST models including fine-tuning, i.e., \vort{MNI}, \vstd{MNI}, and \vrrPow{MNI}{}, results are shown in \myreffig{fig:MNacc} for 5 repeated runs. 
We can see that \vrrPow{MNI}{} outperforms \vort{MNI} and \vstd{MNI}, which indicates that racecar training yields generic features that can also improve performance for the original task. 

As generic features are typically more robust than specific features
\cite{novak2018sensitivity}, we investigate 
a perturbed data set with motion blur (n-MNIST \cite{basu2017learning})
for the task transfer. 
We fine-tune all three previous models for the n-MNIST data set,
to obtain the models \vort{n-MNI}, \vstd{n-MNI}, and \vrrPow{n-MNI}{}. 
Performance results are likewise given in \myreffig{fig:MNacc}. 
Based on the same CNN architecture and parameters, \vrrPow{$mni$}{} achieves the best performance. 
This indicates that the \vrrPow{}{} model learned more generic features 
via racecar training than \vort{} and \vstd{}. Interestingly,
the racecar training not only improves inference accuracy, but also stabilizes
training, as indicated by the variance of the results in \myreffig{fig:MNacc}. 

\begin{figure}[t!]
	\begin{center}
		\includegraphics[width=0.95\linewidth]{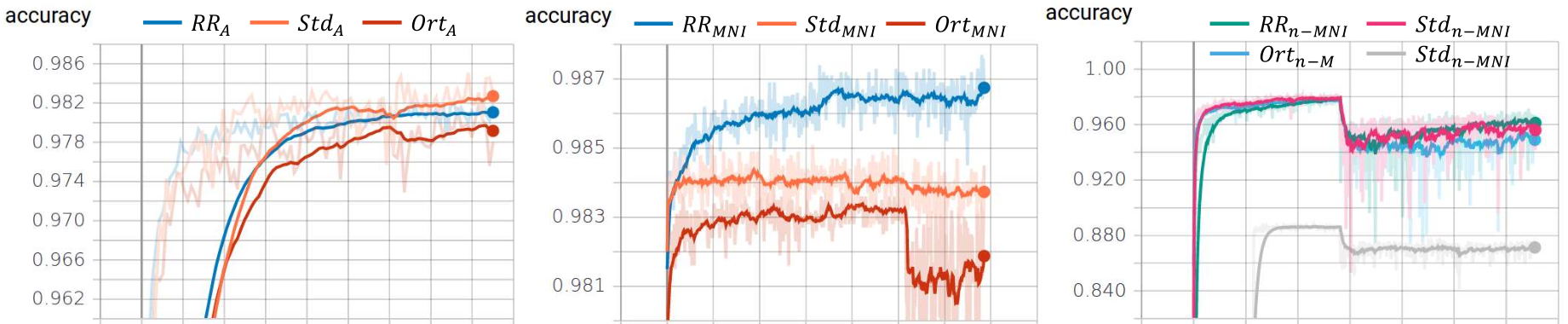}
	\end{center}
	\caption{\footnotesize{
        Visualizations of training processes for the MNIST cases. 
        Left, constraints enabled: \vrrPow{$mni$}{} (accuracy: 0.9810, perf.: 5.675 s/epoch), \vstd{$mni$} (accuracy: 0.9827, perf.: 3.522 s/epoch) and \vort{$mni$} (accuracy: 0.9792, perf.: 4.969 s/epoch). 
        Middle, fine tuning: \vrrPow{MNI}{}, \vstd{MNI} and \vort{MNI}. 
        Right, task transfer: \vrrPow{n-MNI}{}, \vstd{n-MNI}, \vort{n-MNI}, and \vstd{B}.
        }
	}
	\label{fig:MNIST}
\end{figure}

For fine-tuning the models without regularization, i.e.
for training \vrrPow{MNI}{}/\vrrPow{n-MNI}{}, \vstd{MNI}/\vstd{n-MNI} and \vort{MNI}/\vort{n-MNI},
we load \vrrPow{$mni$}{}, \vstd{$mni$}, and \vort{$mni$}, and continue training without any orthogonalization or racecar loss terms.
Example training processes of the MNIST tests are shown in \myreffig{fig:MNIST}. We can see that the racecar loss increases the task difficulty, so \vrrPow{$mni$}{} exhibits a slightly lower performance and longer training time than \vstd{$mni$} in the first phase. In the second phase \vrrPow{$mni$}{} outperforms \vstd{$mni$} and \vort{$mni$} for both tasks. 
This illustrates that the dominant features extracted by racecar training 
also improved performance for both task domains.
Numeric accuracies of all models are given in Tab. 3 and Tab. 4.

\subsection{Natural Image Classification} 

\label{sec:Cifartest}
\begin{figure}
\begin{minipage}{.49\linewidth}
    \centering
        \includegraphics[height=1.7in]{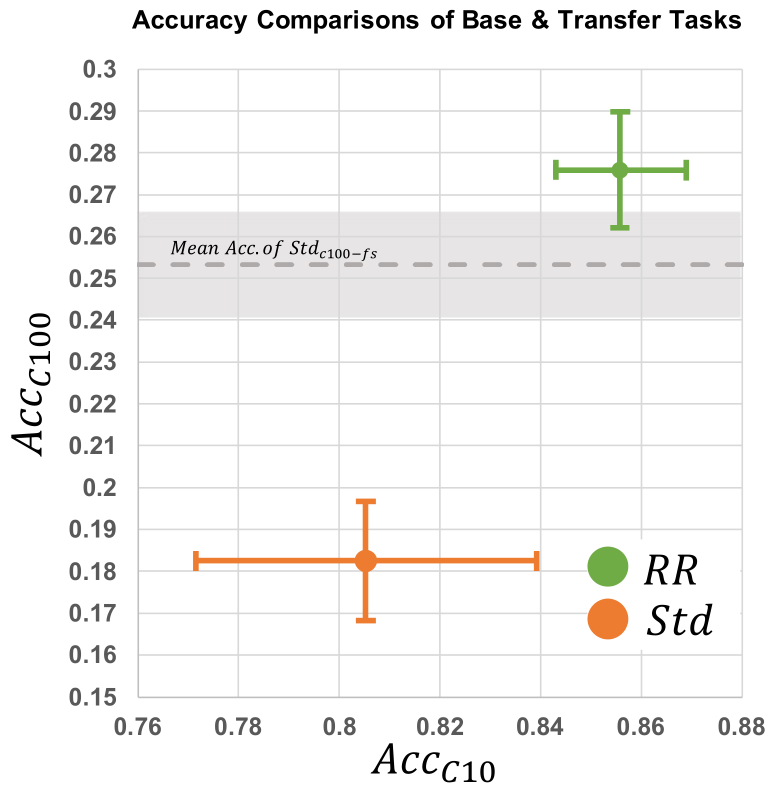}
        \captionsetup{width=.9\textwidth}
    \caption{\footnotesize{
        Accuracy comparisons of natural image tasks $A$ and $B$. 
        Grey line and grey region represent accuracy of baseline \vstd{C100-fs}. 
        \vrrPow{C10}{} and \vrrPow{C100}{} got best performance for task $A$ and $B$.}
    } \label{fig:CIacc}
\end{minipage}
\begin{minipage}{.49\linewidth}
\centering
    \includegraphics[height=1.7in]{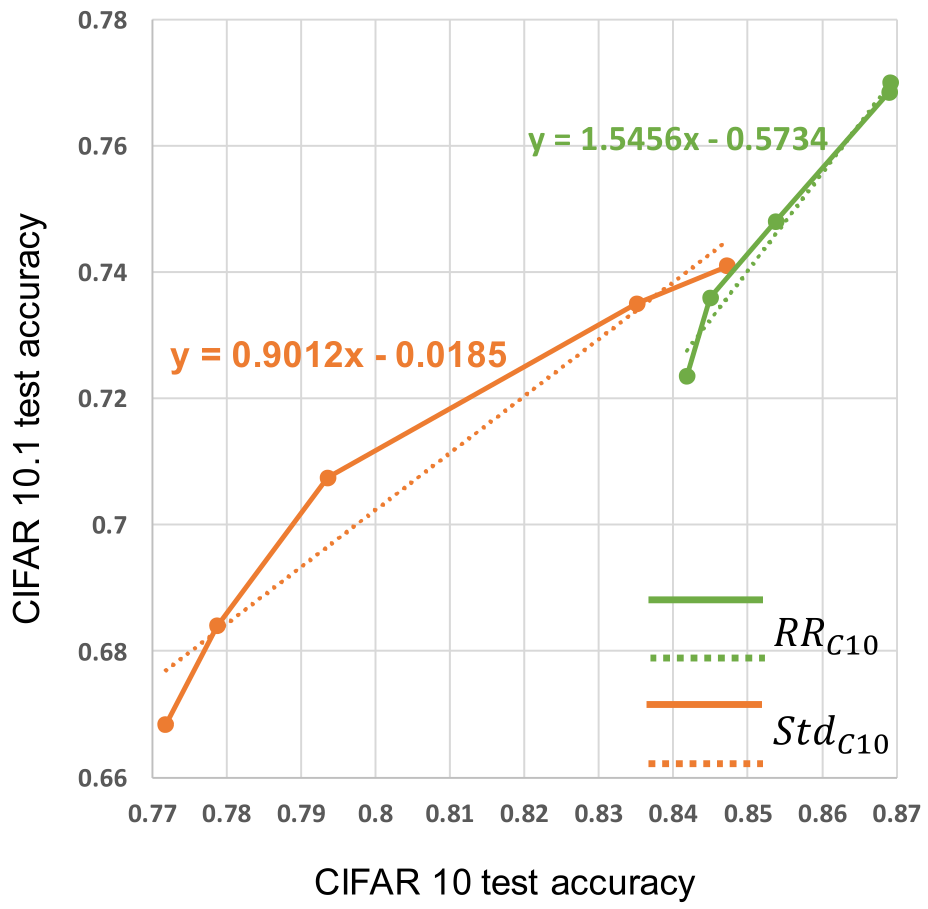}
    \vspace{-2mm}
    \captionsetup{width=.9\textwidth}
    \caption{\footnotesize{
    Accuracy comparisons when applying models trained on CIFAR 10 to CIFAR 10.1 data. }}
    \label{fig:cifar10_10.1}
\end{minipage}
\end{figure}
Natural images arise in many important application scenarios, 
and in Sec.5 of the main paper we evaluated racecar training for a very deep ResNet110
for CIFAR 10 \cite{krizhevsky2009learning}.
This experiment already demonstrated that racecar training can yield substantial gains in 
SOTA performance compared to orthogonality constraints.
These gains are on a level that is similar with gains that orthogonality constraints 
exhibit over a regular training.
Details of the corresponding Resnet110 accuracies, network architecture, and hyperparameters are shown in Tab. 7, 17 and 18, respectively. To show that racecar training also yields gains in shallow networks, we 
additionally evaluate our approach with the CIFAR data set using a 19-layer network.

\paragraph{Training Details:} Both CIFAR 10 and CIFAR 100 data sets consist of $60k$ $32\times32$ images. $50k$ of them are used for training and $10k$ of them are used for testing. 
classes. The forward and reverse pass architectures of the network can be found in Tab. 19 with parameters given in Tab. 20. Models are trained using cross-entropy as loss function, and
all 13 convolutional layers are used for the racecar loss. 
For fine-tuning on CIFAR 100, the last layer is omitted when loading due to the changed size of the output.

\begin{figure}[t!]
	\begin{center}
		\includegraphics[width=0.99\linewidth]{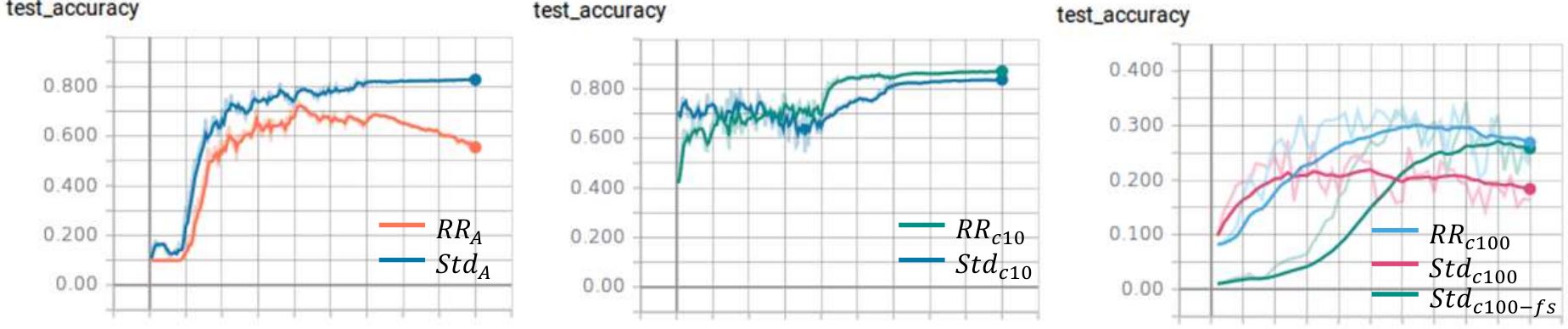}
	\end{center}
	\caption{\footnotesize{
        Visualizations of training processes. 
        Left: CIFAR 10 with constraints, \vrrPow{c1}{} (accuracy: 0.5784, cost: 64 seconds/epoch) and \vstd{c1} (accuracy: 0.8272, cost: 63 seconds/epoch). 
        Middle: CIFAR 10 , fine tuning, \vrrPow{C10}{} and \vstd{C10}. 
        Right: Task transfer for CIFAR 100, \vrrPow{C100}{}, \vstd{C100} and \vstd{C100-fs}.
        }
	}
	\label{fig:Cifar}
\end{figure}

\paragraph{Analysis of Results:} 
At first, we train two models with the CIFAR 10 data set as classification task $C10$, i.e., \vrrPow{c1}{} and \vstd{c1}. Example training processes of the CIFAR tests are shown in \myreffig{fig:Cifar}. 
We fine-tune to obtain \vrrPow{C10}{} and \vstd{C10}, results for which are shown in \myreffig{fig:CIacc}. Numerical results are given in Tab. 5. The racecar training also improves performance for this natural image classification task.
As transfer learning task we fine-tune \vrrPow{c1}{} and \vstd{c1} for the CIFAR 100 data set classification, yielding models \vrrPow{C100}{} and \vstd{C100}. Results are likewise shown in \myreffig{fig:CIacc}.
Training with the same CNN architecture and parameters, 
\vstd{c1} has difficulties adjusting to the new task, while our model from the initial racecar training even slightly outperforms a model trained from scratch for CIFAR 100 (\vstd{C100-fs}).

As in the main text, we evaluate these models with the CIFAR transfer benchmark from Recht et al. \cite{recht2019imagenet}. It measures how well models trained on the original CIFAR data set, i.e. CIFAR 10, generalize to a new data set (CIFAR 10.1) collected according to the same principles as the original CIFAR 10. 
Here, it is additionally interesting to see how well performance for CIFAR 10 translates into transfer performance for 10.1. 
As shown in \myreffig{fig:cifar10_10.1},
estimating the slope of this relationship with a linear fit yields $0.9012$ for \vstd{C10}, and $1.5456$ for \vrrPow{C10}{}. Thus, while regular training performs worse on the new data when its performance for the original task increases (with a slope of less than one), the models 
from racecar training very successfully translate gains in performance from the original task to the new one. 
This indicates that the models have successfully learned a set of generalizing features. 

\section{Experimental Results}
\label{sec:results}

Below, we give additional details for the experiments of Sec. 5 of the main paper.

\subsection{Texture-shape Benchmark}
\label{sec:Tstestapp}

\paragraph{Training Details:} 
All training data of the texture-shape tests were obtained from \cite{geirhos2018imagenet}. The stylized data set contains 1280 images, 1120 images are used as training data, and 160 as test data. Both edge and filled data sets contain 160 images each, all of which are used for testing only. All three sets (stylized, edge, and filled) contain data for 16 different classes. The forward and reverse pass architectures of the network are given in Tab. 13, with hyperparameters in Tab. 14. Numerical accuracy for \vrrPow{TS}{}, \vort{TS} and \vstd{TS} is given in Tab. 27, Tab. 28 and Tab. 29, respectively. 
\begin{wrapfigure}{hR}{0.55\linewidth}
    \centering
        \includegraphics[width=0.99\linewidth]{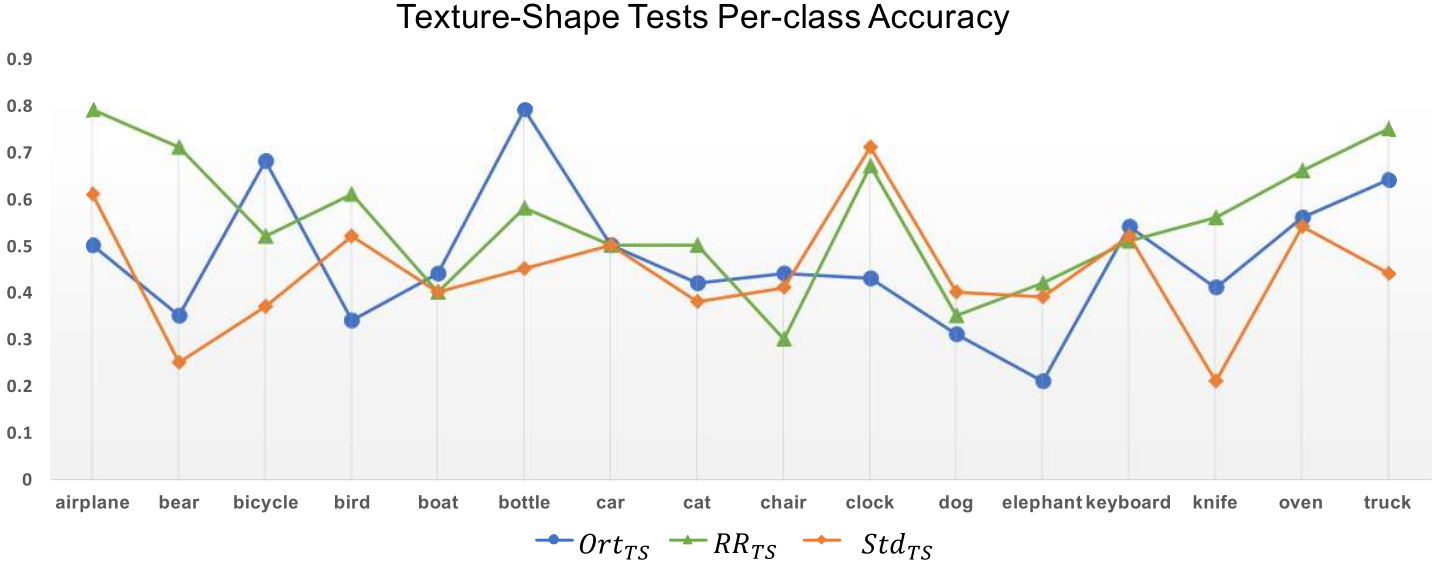}
    \vspace{-3mm}
    \caption{\footnotesize{Test accuracy for all 16 classes for the three model variants.}
    } \label{fig:texture_all}
\vspace{-4mm}
\end{wrapfigure}
\paragraph{Analysis of Results}:
For a detailed comparison, we list per-class accuracy of stylized data training runs for \vort{TS}, \vstd{TS} and \vrrPow{TS}{} in
\myreffig{fig:texture_all}. \vrrPow{TS}{} outperforms the other two models for most of the classes.
Training times (in seconds) for different models are listed in Tab. 30. \vrrPow{TS}{} requires $41.86\%$ more time for training compared to \vstd{TS}, but yields a $23.76\%$ higher performance. 
All models saturated, i.e. training \vstd{TS} or \vort{TS} longer does not increase classification accuracy any further.
We also investigated how much we can reduce model size when using racecar training in comparison to the baselines.
A reduced racecar model 
only has $67.94\%$ of the parameters, while still outperforming \vort{TS}.

\subsection{Generative Adversarial Models}
\label{sec:Smoketestapp}

\paragraph{Training Details:} 
The data set of smoke simulation was generated with a fluid solver from 
an open-source library \cite{mantaflow}. 
We generated 20 simulations with 120 frames each, with 10\% of the data being used for training. Smoke inflow region, inflow velocity, and buoyancy force were randomized to produce varied data. The low-resolution data were down-sampled from the high-resolution data by a factor of 4. Data augmentation, such as flipping and rotation was used in addition.
As outlined in the main text, we consider building an autoencoder model for the synthetic data as task $B_1$, 
and a generating samples from a real-world smoke data set as task $B_2$. 
The smoke capture data set for $B_2$ contains 2500 smoke images from the ScalarFlow data set \cite{eckert2019scalarflow}, 
and we again used 10\% of these images as training data set. 

Task $A$: 
We use a fully convolutional CNN-based architecture for generator and discriminator networks (cf. Tab. 21, and parameters in Tab. 23.).
Note that the inputs of the discriminator contain high resolution data $(64,64,1)$, as well as low resolution $(16,16,1)$, which is up-sampled to $(64,64,1)$ and concatenated with the high resolution data.
In line with previous work \cite{xie2018tempogan}, \vrrPow{A}{} and \vstd{A} are trained with a non-saturating GAN loss, feature space loss and L2 loss as base loss function. All generator layers are involved in the racecar loss.

Task $B_{1}$:
All encoder layers are initialized from \vrrPow{A}{} and \vstd{A} when training \vrrPow{$AB_{1}$}{} and \vstd{$AB_{1}$}. It is worth noting that the reverse pass of the generator is also constrained when training \vrrPow{A}{}. So both encoder and decoder are initialized with parameters from \vrrPow{A}{} when training \vrrPow{$AB_{1}$}{}.
This is not possible for a regular network like \vstd{$AB_{1}$}, as the weights obtained with a normal 
training run are not suitable to be transposed. Hence, the de-convolutions of \vstd{$AB_{1}$}
are initialized randomly. 

Task $B_{2}$:
As the data set for the task $B_{2}$ is substantially different and contains RBG images 
(instead of single channel gray-scale images), 
we choose the following setups for the \vrrPow{A}{} and \vstd{A} models: 
parameters from all six layers of \vstd{A} and \vrrPow{A}{} are reused for initializing 
decoder part of \vstd{$AB_{2}$} and \vrrPow{$AB_{2}$}{}. Specially, when initializing the last layer of of \vstd{$AB_{2}$} and \vrrPow{$AB_{2}$}{}, we copy and stack the parameters from the last layer of \vstd{A} and \vrrPow{A}{} into three channels to fit output data size in task $B_{2}$. Here, the encoder part of \vrrPow{$AB_{2}$}{} is not initialized with \vrrPow{A}{}, due
to the significant gap between training data sets of task $B_{1}$ and task $B_{2}$. Our experiments show that only initializing the decoder part of \vrrPow{$AB_{2}$}{} (avg. loss:$1.56e7$, std. dev.:$3.81e5$) outperforms initializing both encoder and decoder (avg. loss:$1.82e7$, std. dev.:$2.07e6$). We believe the reason is that initializing both encoder and decoder part makes it more difficult to adjust the parameters for new data set that is very different from the data set of the source task.

\begin{figure}[t!]
	\begin{center}
		\includegraphics[width=0.99\linewidth]{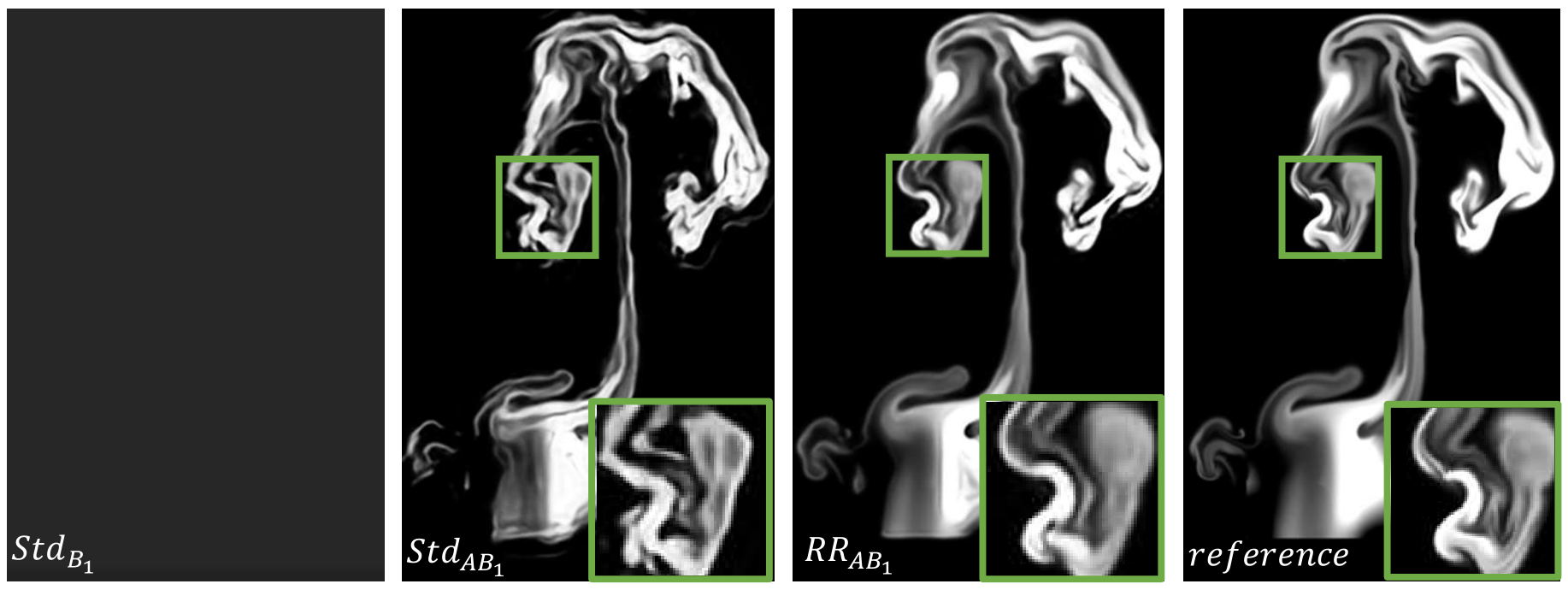}
	\end{center}
	\caption{\footnotesize{Example outputs for \vrrPow{$AB_{1}$}{}, \vstd{$AB_{1}$}, \vstd{$B_{1}$}. The reference is shown for comparison. \vrrPow{$AB_{1}$}{} produces higher quality results than \vstd{$AB_{1}$}, while a model trained from scratch, \vstd{$B_{1}$} fails for this task. It produces a mostly black image. }}
	\label{fig:B1output}
\end{figure}

\begin{figure*}[t!]
    \centering
    \includegraphics[width=0.99\linewidth]{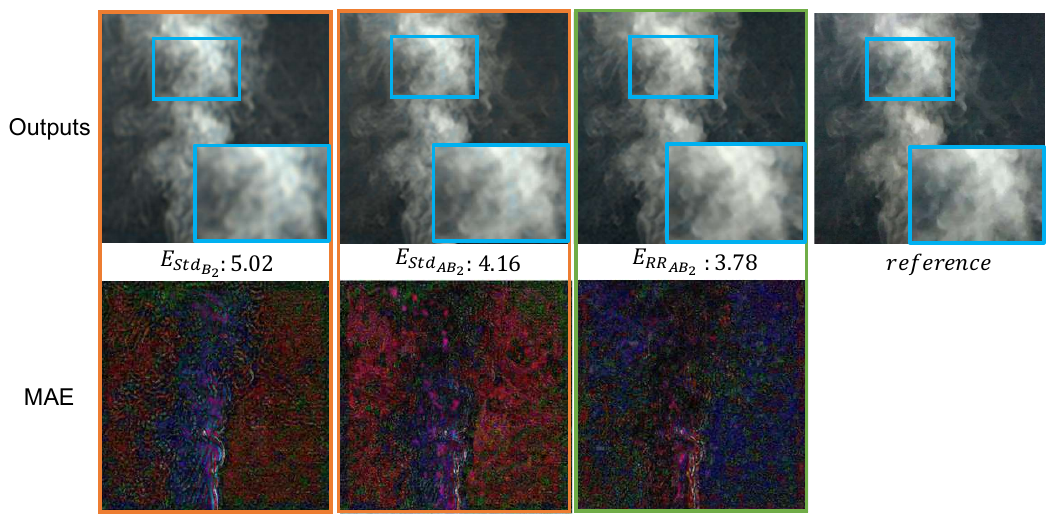}
    \caption{\footnotesize{Mean Absolute Error (MAE) comparisons for smoke task $B_2$ models. \vrrPow{$AB_{2}$}{} shows the smallest error, and additionally achieves the best visual quality amongst the different models. It even outperforms the baseline model trained from scratch, i.e. \vstd{$B_2$}.
    }
    } \label{fig:capAEdiff}
\end{figure*}

\paragraph{Analysis of Results:} 
We first discuss the autoencoder transfer learning task $AB_{1}$ for synthetic, i.e., simulated fluid data.
(architectures for $B_{1}$ and $B_{2}$ are given in Tab. 22).
Example outputs of \vrrPow{$AB_{1}$}{}, \vstd{$AB_{1}$} and \vstd{$B_{1}$} are shown in ~\myreffig{fig:B1output}. It is clear that \vrrPow{$AB_{1}$}{} gives the best performance among these models.
We similarly illustrate the behavior of the transfer learning task $AB_{2}$ for images of real-world fluids. This example likewise uses an autoencoder structure. Visual comparisons are provided in \myreffig{fig:capAEdiff}, where  \vrrPow{$AB_{2}$}{} generates results that are closer to the reference.
Overall, these results demontrate the benefits of racecar training 
for GANs, and indicate its potential to obtain more generic
features from synthetic data sets that can be used for tasks involving 
real-world data. 

\subsection{VGG19 Stylization}
\label{sec:Vggtestapp}
The section below gives details of the stylization tests in Sec. 5.4 of the main paper.

\paragraph{VGG19 Training:}  
For the ImageNet data set \cite{deng2009imagenet}, 1281167 images of 1000 classes are used for training, and 50k images are used for testing. The size of all images is $224\times224$. The forward and reverse pass of VGG19 are given in Tab. 24, with hyperparameters shown in Tab. 25.
All 16 convolutional layers are used for the racecar loss. To speed up the training process of \vrrPow{}{}, we first train a model without racecar loss for 6 epochs with batch size 64. We then reuse this model for training \vrrPow{}{} and \vstd{} with a batch size of 24 for 30 epochs. 

\paragraph{Stylization Background:} 
Gatys et al. \cite{gatys2016image} propose to use a loss consisting 
of a content and a style term, i.e. $\Loss{\text{total}}={\eta}\Loss{\text{content}} + {\delta}\Loss{\text{style}}$, 
 where $\eta$ and $\delta$ denote coefficients. $\Loss{\text{content}}$ calculates a content difference between source image $p$ and generated image $g$: 
\begin{equation}
\Loss{\text{content}}(p, g, t)=\frac{1}{2}\sum_{m,n}(F^{m,n,t}_{p}-F^{m,n,t}_{g})^2,
\label{eq:contentloss}
\end{equation}
where $F^{m,n,t}_{p}$ is the feature representation of $p$ of the $m^{th}$ filter at position $n$ in layer $t$.
$\Loss{\text{style}}$ instead computes a style difference between style image $a$ and a generated image $g$:
\begin{equation}
             \begin{array}{lr}  
             G_{a}^{m,n,t}=\sum_{f}F_{a}^{m,f,t}F_{a}^{n,f,t},  \\  
             E_{t}=\frac{1}{4N^2_{t}M^2_{t}}\sum_{m,n}(G_{g}^{m,n,t}-G_{a}^{m,n,t})^{2}, \\  
             \Loss{\text{style}}(a, g)=\sum_{t=0}^{T}\omega _{t}E_{t},
             \end{array}  
\label{eq:styleloss}
\end{equation}  
where $\omega _{t}$ is are the weighting factors for layer $t$; 
$G$ is the Gram matrix; $N_{t}$ denotes filter numbers of layer $t$, 
and $M_{t}$ is the dimension of layer $t$'s filter. $T$ denotes 
the  number of layers included in the style loss.

\subsection{Additional Stylization Results}
\label{sec:addStyle}

\begin{wrapfigure}{hR}{0.4\linewidth}
    \centering
        \includegraphics[width=0.99\linewidth]{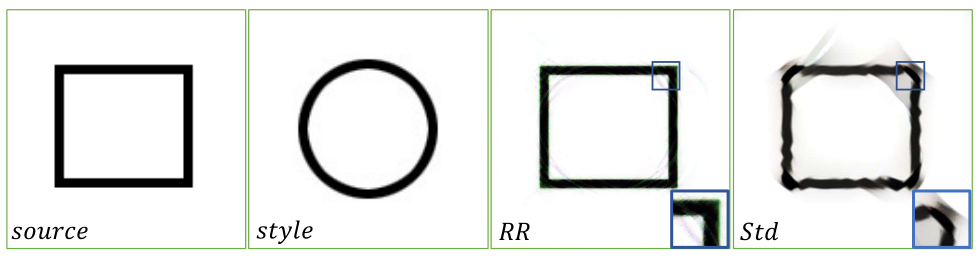}
    \caption{\footnotesize{\nilsE{
        Primary shape stylization test: \vrrPow{}{} manages to preserve the sharp 
        edges and clear lines of the source.
    }}
    } \label{fig:vgg1}
\end{wrapfigure}

To compare the feature extracting capabilities between the \vrrPow{}{} and 
\vstd{} runs, we only optimize $\Loss{\text{style}}$ as a first test.
Two geometric shapes are given as source and style images,
as shown in \myreffig{fig:vgg1}. 
They have the same background and object color. The only difference is the large-scale shape of the object. Thus, as we aim for applying a style that is identical to the source, we can test whether the features can clearly separate and preserve the large scale shape features from the ones for the localized style.
Comparing the results of \vrrPow{}{} and \vstd{}, 
the output of \vrrPow{}{} is almost identical to the input,
while stylization with \vstd{} changes the shape of the object
introducing undesirable streaks around the outline. Thus, \vrrPow{}{} performs significantly better.

\begin{figure}[t!]
\begin{minipage}{.49\linewidth}
  \centering
  \includegraphics[height=1.7in]{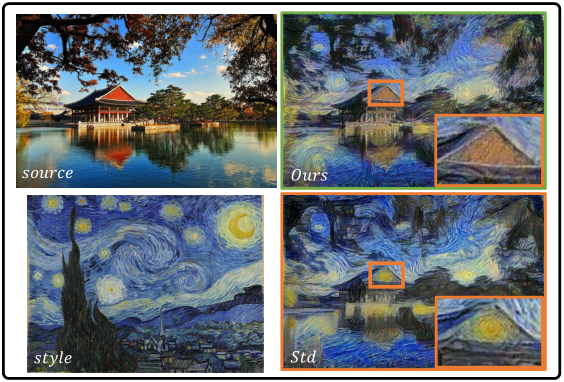}
  \captionsetup{width=.94\textwidth}
  \caption{\footnotesize{Stylization (natural image to van Gogh).}}
  \label{fig:starvan}
\end{minipage}
\begin{minipage}{.49\linewidth}
  \centering
  \includegraphics[height=1.7in]{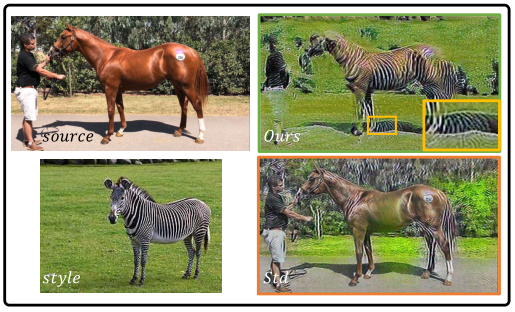}
  \captionsetup{width=.94\textwidth}
  \caption{\footnotesize{Stylization (horse to zebra).}}
  \label{fig:starhorse}
\end{minipage}
\end{figure}

\myreffig{fig:starvan} shows a painterly style transfer, where both $\Loss{\text{style}}$ and $\Loss{\text{content}}$ are optimized. Here, the regular \vstd{} model mixes sky and house structures, indicating that the extracted features fail to separate style and content information.
A more difficult case with a horse-to-zebra transfer is shown in \myreffig{fig:starhorse}. After optimizing $\Loss{\text{content}}$ and $\Loss{\text{style}}$, our method generates a zebra pattern in the horse's shadow (zoom-in in the yellow box), but yields improved results compared to a regular model.
Those results indicate that \vrrPow{}{} is able to clearly 
encode the shape of the object and preserve it during the stylization,
while the features of \vstd{} fail to clearly separate shape and style.

 \begin{figure}[t!]
     \centering
     \includegraphics[width=0.99\linewidth]{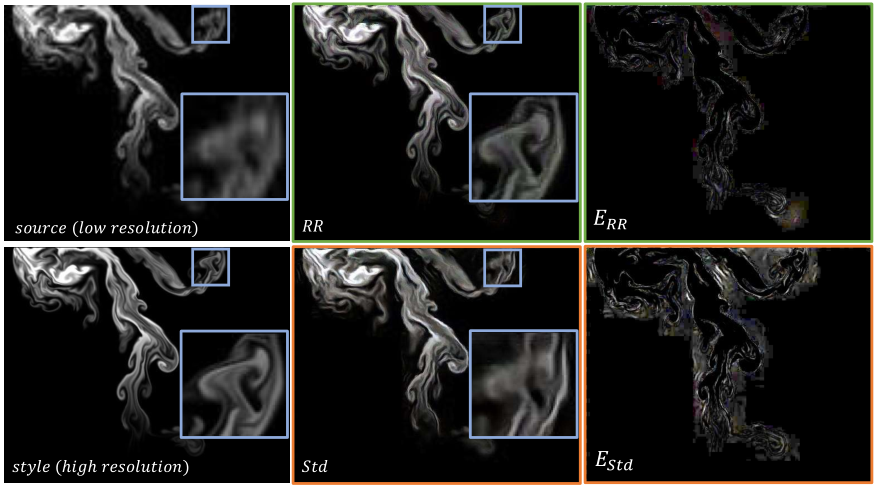}
     \caption{\footnotesize{Comparison of stylization from low- to high-resolution simulation. Results of \vrrPow{}{} are closer to the high resolution reference.}
     } \label{fig:vgg4}
     \vspace{-2mm}
 \end{figure}

 We additionally confirm the capabilities of the VGG model from racecar training with 
 a low- to high-resolution transfer where the goal is purely to add detail.
 I.e., we can compute in-place errors w.r.t. a high-resolution reference.
 Both $\Loss{\text{content}}$ and $\Loss{\text{style}}$ are optimized and the result is shown in \myreffig{fig:vgg4}. 
The zoomed-in areas in the figure highlight these differences.
The VGG model obtained with a regular training run yields significantly higher errors than the racecar model, 
as shown on the right side of \myreffig{fig:vgg4}. Here, $E_{\text{Std}}$ contains substantially 
larger errors w.r.t. the reference than $E_{\text{RR}}$.

 \begin{figure*}[t!]
     \centering
     \includegraphics[width=0.85\linewidth]{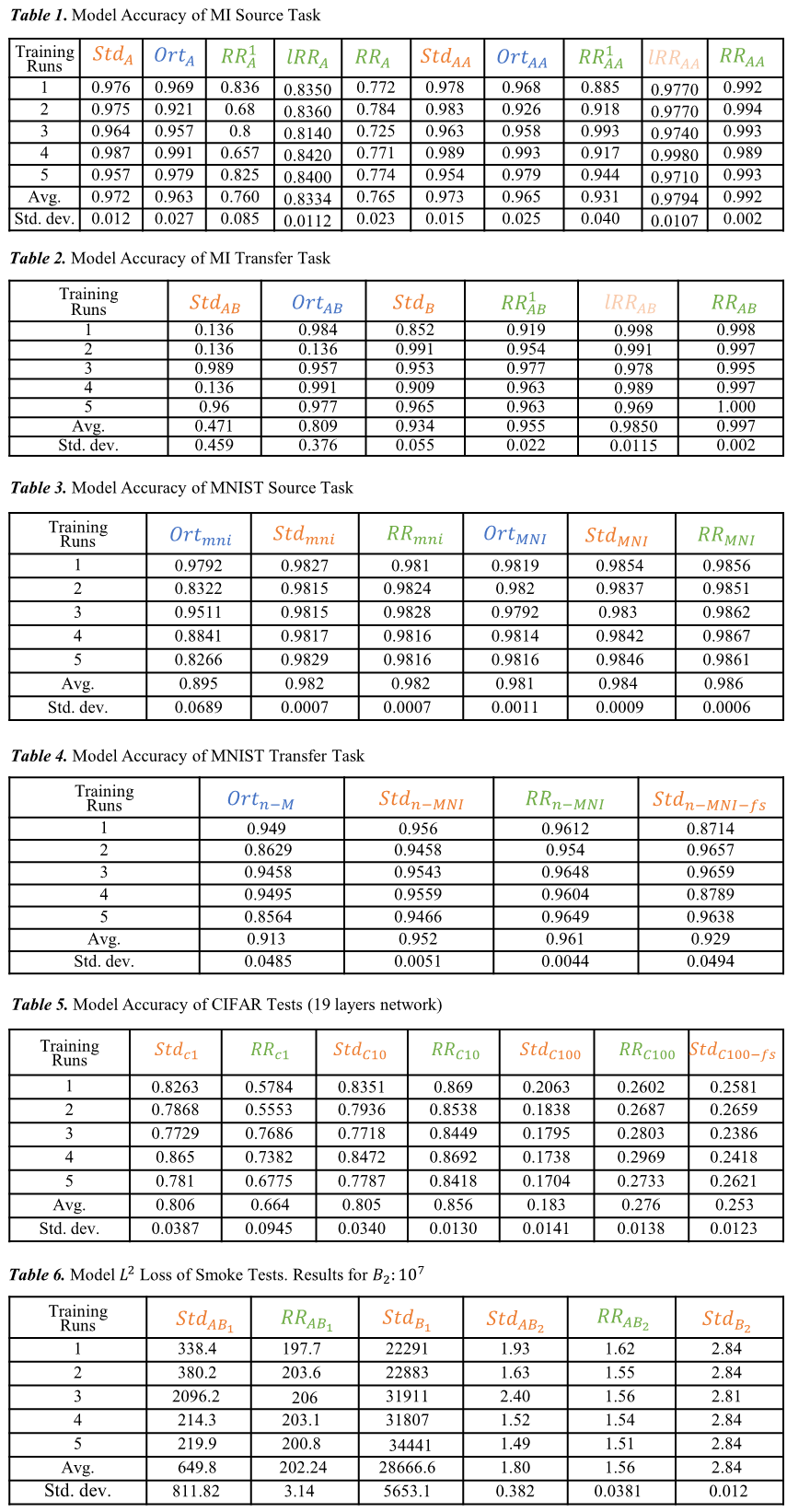}
 \end{figure*}

  \begin{figure*}[t!]
     \centering
     \includegraphics[width=0.9\linewidth]{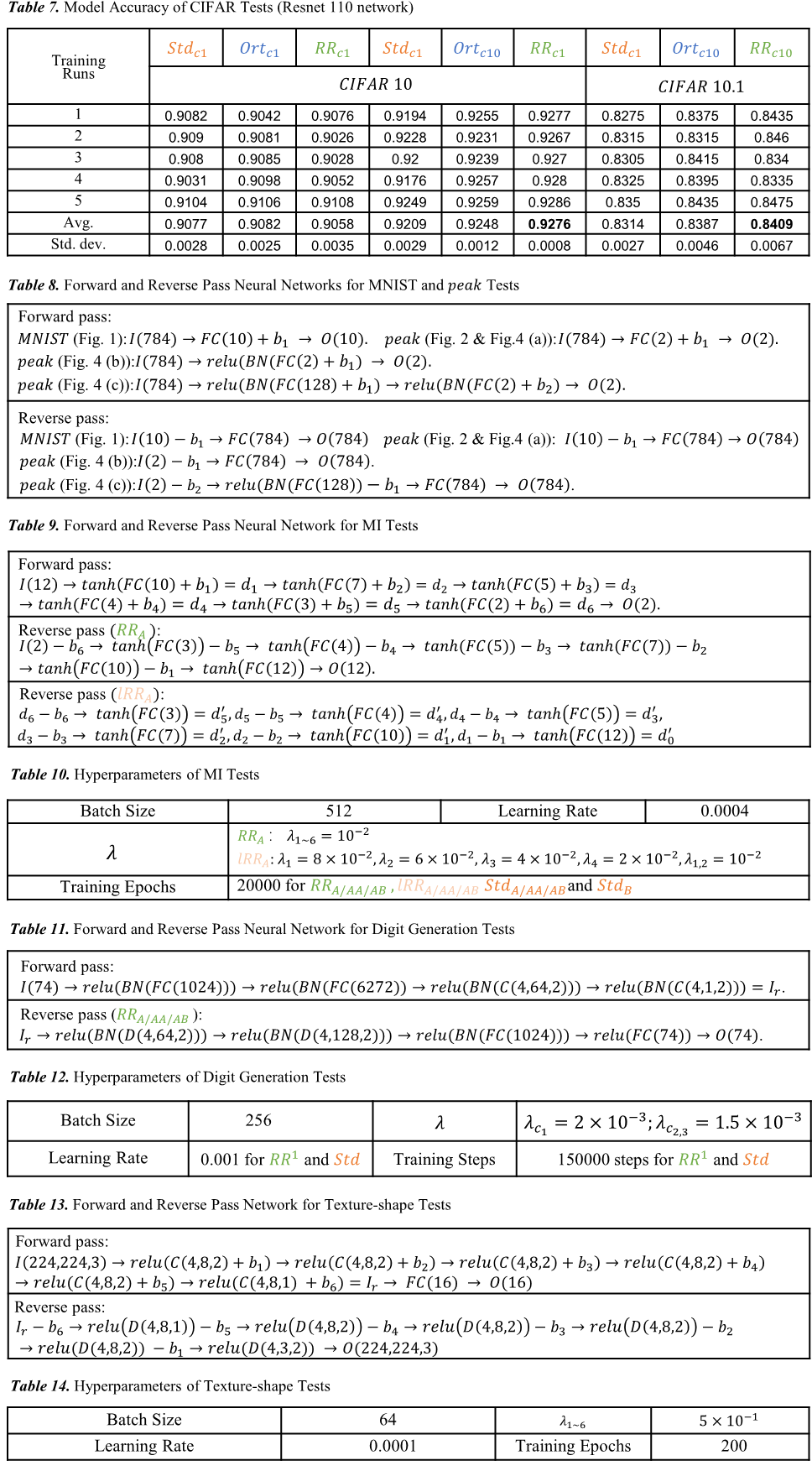}
 \end{figure*}
 
  \begin{figure*}[t!]
     \centering
     \includegraphics[width=0.9\linewidth]{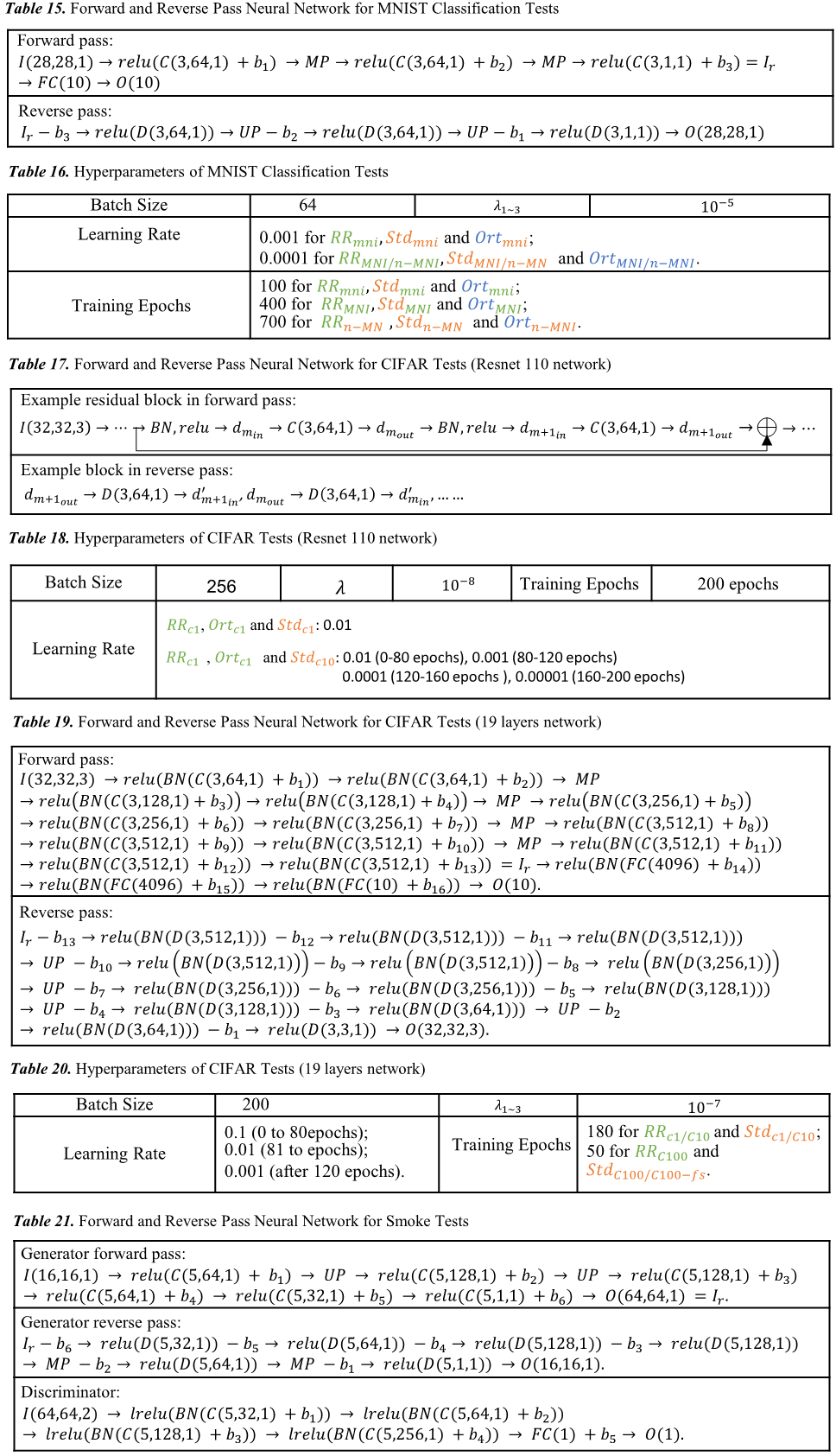}
 \end{figure*}
 
  \begin{figure*}[t!]
     \centering
     \includegraphics[width=0.92\linewidth]{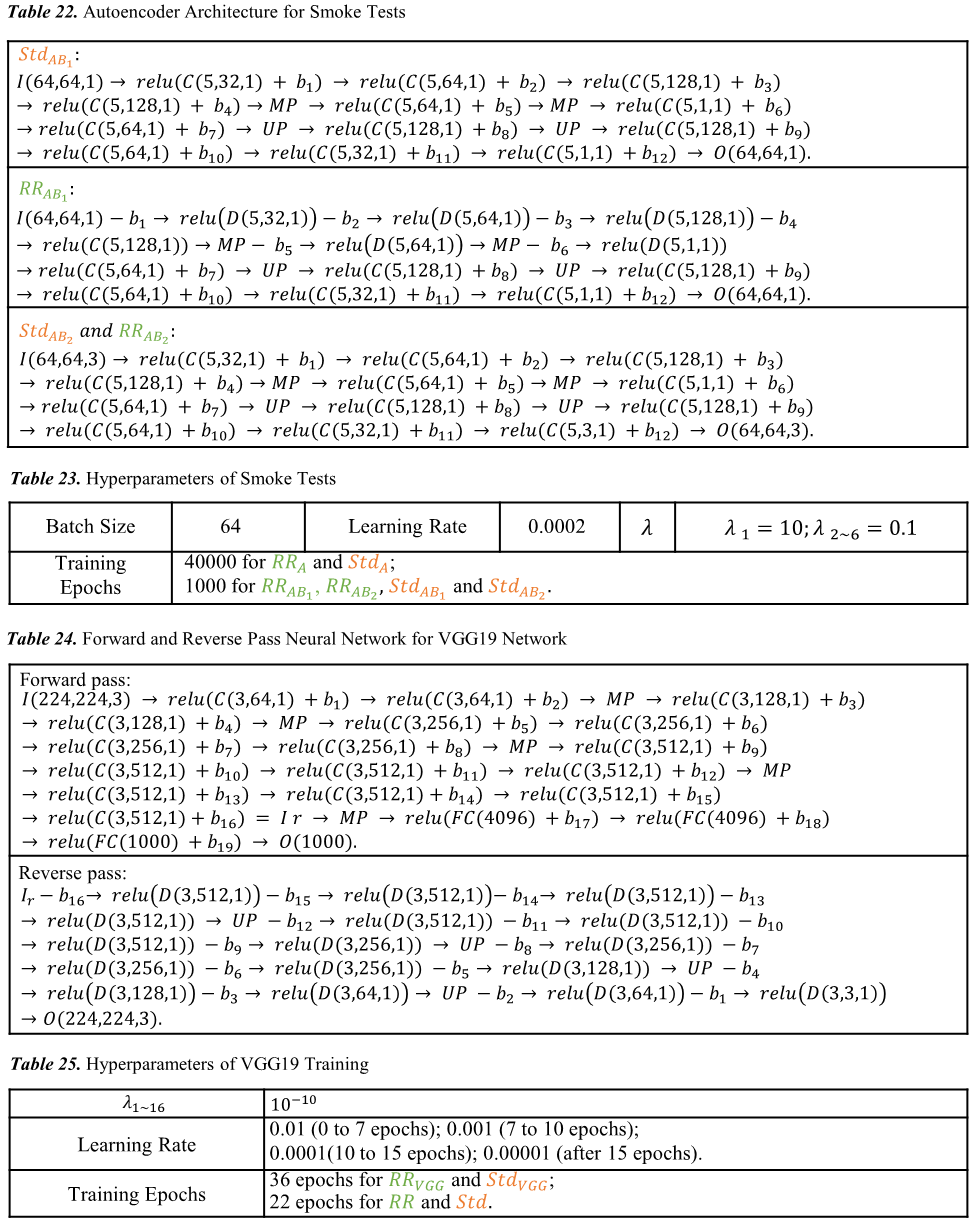}
 \end{figure*}

  \begin{figure*}[t!]
     \centering
     \includegraphics[width=0.99\linewidth]{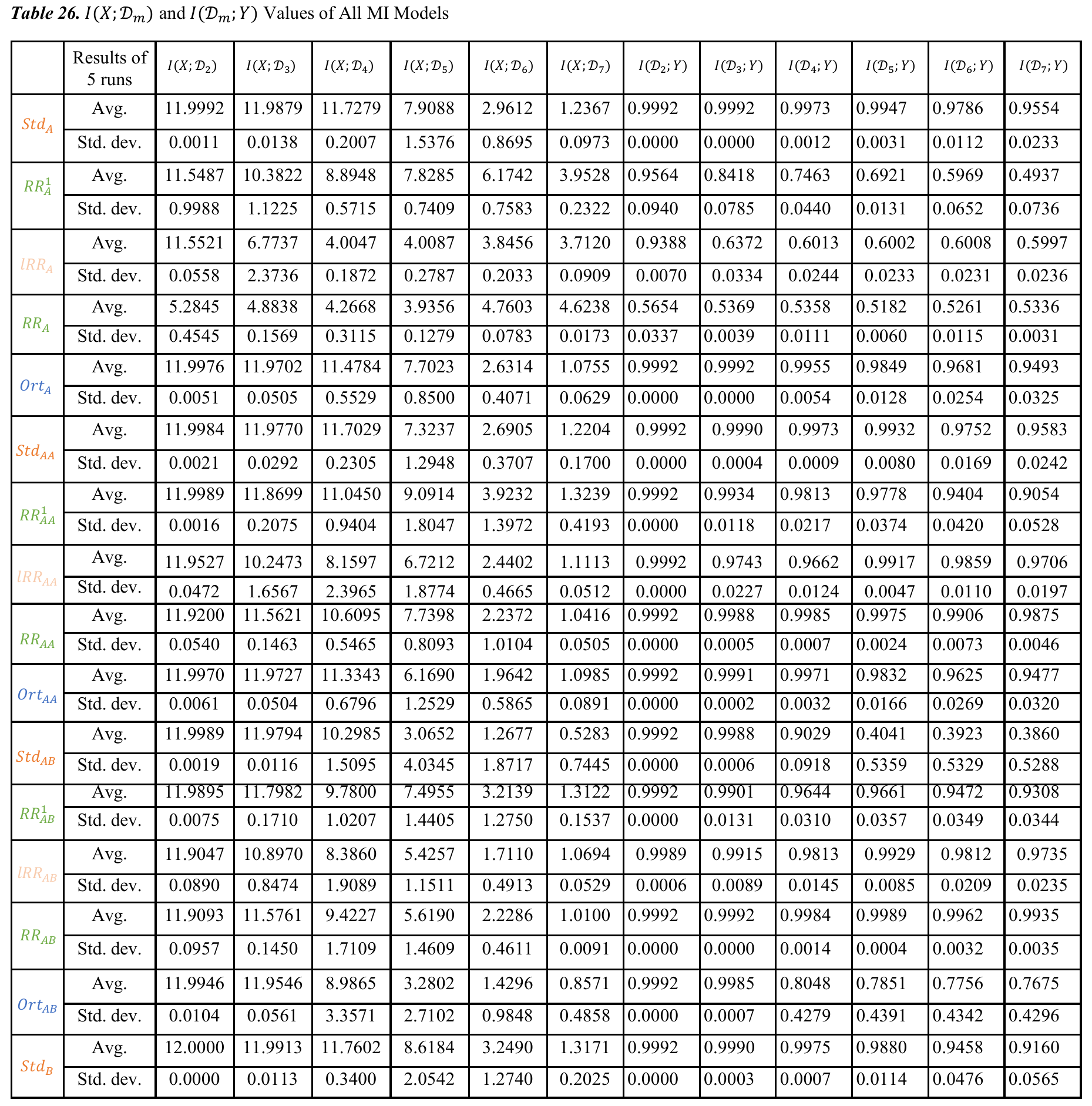}
 \end{figure*}
 
   \begin{figure*}[t!]
     \centering
     \includegraphics[width=0.99\linewidth]{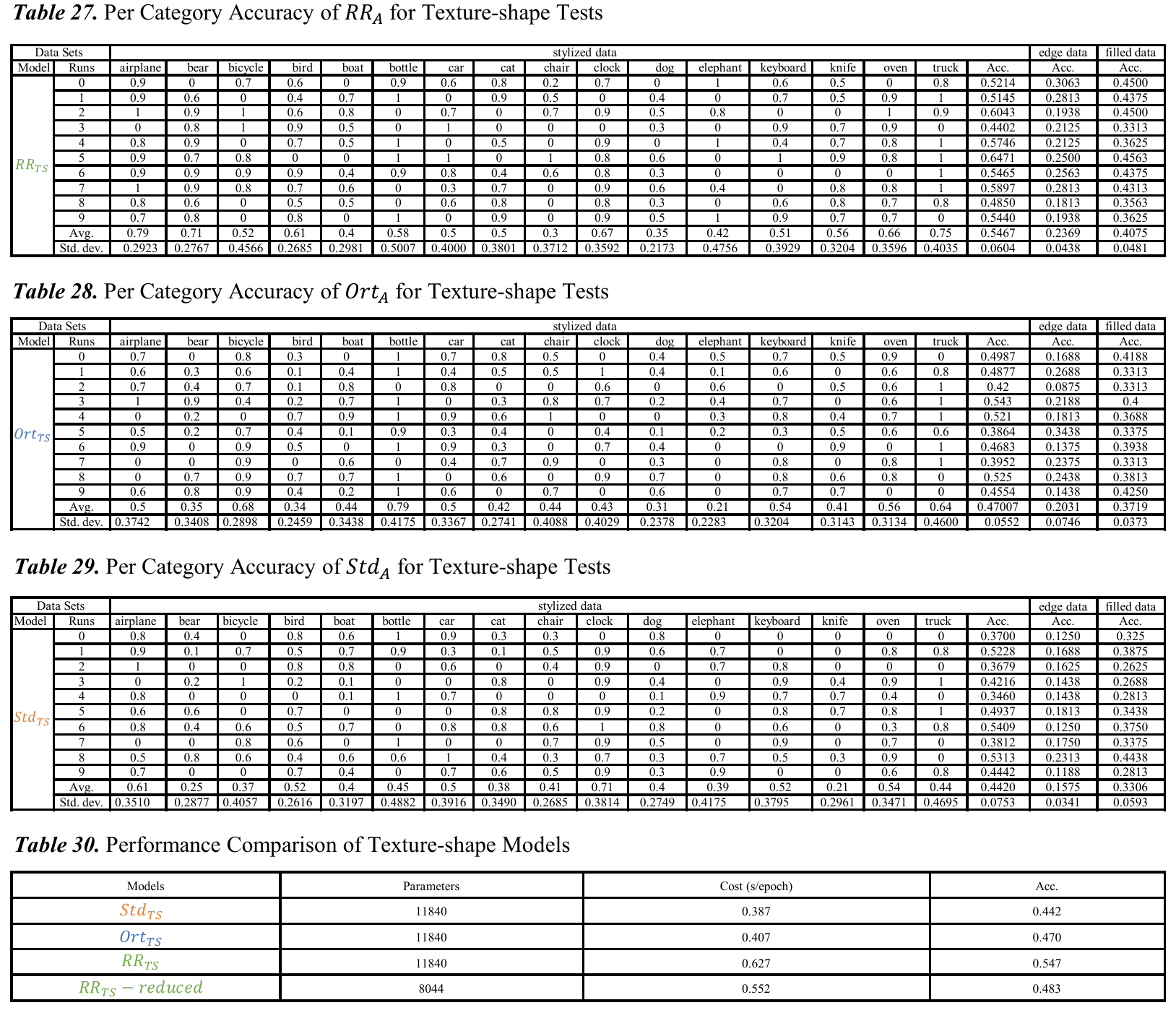}
 \end{figure*}

\section{Impact Statement}

Racecar training could be applied to a wide range of applications, such as computer vision, computer graphics, classification tasks, generative tasks, and many more. We specifically focused on transfer learning in our paper. 
\nilsE{For regular training runs, models are trained to establish a direct relationship between the input and output distribution via a specific loss function. We instead aimed at learning dominant features from the dataset via building a forward and reverse pass network. Motivated from the way humans learn, it improves the transferability of trained models. 
Generally, if we use the same data set for different purposes, we have to train different models from scratch. With racecar training, we can train a model which contains the dominant features of a data set as a starting point, and then reuse this model for different purposes. This is a more efficient process that likewise could be beneficial for a wide range of applications. 

Besides, our research could also be used to analyze whhich input features are extracted by a layer, 
and thus improve the interpretability of neural networks in general. 
Hence, racecar training can be applied when the dominant features or modes of a data set are essential for making decisions about how to employ learned models.}

Those properties could also result in potential risks: e.g., for data sets with personal data, i.e., in situations where privacy is critical, a learned representation using racecar training might contain more sensitive personal details than when performing a regular training. Hence, the acquired models will require additional safety measures. 
Thus, racecar training is, at the same time, a potential tool for analyzing the safety of a learned representation, as well as a potential risk.
We encourage researchers to consider these aspects and risks arising from them for future directions of their work.

\bibliography{neurips_2020}
\bibliographystyle{plain}

\end{document}